\PassOptionsToPackage{table}{xcolor}
\documentclass{article}

\PassOptionsToPackage{numbers, compress}{natbib}

\usepackage[preprint]{mystyle}
\usepackage{amsmath}
\usepackage[utf8]{inputenc}
\usepackage[T1]{fontenc}
\usepackage{hyperref}
\usepackage{url}
\usepackage{booktabs}
\usepackage{amsfonts}
\usepackage{nicefrac}
\usepackage{microtype}

\usepackage{graphicx}
\usepackage{orcidlink}
\usepackage{algorithm, algorithmic}
\usepackage{multirow, tabularx}
\usepackage{bm, threeparttable, upgreek, pifont}
\usepackage{bbding, makecell, wrapfig, soul, wasysym, xspace}
\usepackage{placeins, adjustbox, subcaption}
\usepackage{float}
\usepackage{comment}
\usepackage[normalem]{ulem}

\definecolor{Gray}{gray}{0.92} 
\definecolor{neuripsblue}{rgb}{0.21,0.49,0.74}
\hypersetup{
    breaklinks=true,
    colorlinks=true,
    citecolor=neuripsblue,
    allcolors=neuripsblue
}

\title{TinyFormer: Preserving Tiny Objects in YOLO-DETR Hybrid Real-time Detectors}

\author{%
  Jun-Wei Hsieh\textsuperscript{1}\thanks{*Corresponding author.} \quad
  Meng-Yu Kao\textsuperscript{1} \quad
  Ghufron Wahyu Kurniawan\textsuperscript{1} \quad
  Kuan-Chuan Peng\textsuperscript{2} \\
  \vspace{0.2cm} \\ 
  \textsuperscript{1}College of Artificial Intelligence, National Yang Ming Chiao Tung University, Taiwan \\
  \textsuperscript{2}Mitsubishi Electric Research Laboratories \\
  \texttt{\{jwhsieh, kobekao.ai12, Ghufron.ai13\}@nycu.edu.tw} \quad \texttt{kpeng@merl.com}
}

\def\eg{\emph{e.g.}}

\usepackage{pifont}
\newcommand{\xmark}{\text{\ding{55}}}
\definecolor{checkmark}{HTML}{305AFF}
\definecolor{xmark}{HTML}{E62020}
\newcommand{\ccheck}{{\textcolor{checkmark}{\large\checkmark}}}
\newcommand{\ccross}{{\textcolor{xmark}{\large\xmark}}}

\begin{document}

\maketitle

\begin{abstract}
Both YOLO-series and DETR-based object detectors exhibit systematic limitations in detecting tiny objects. YOLO-style detectors typically rely on backbones with a total downsampling factor of $32\times$, causing tiny objects to vanish at deep feature levels and resulting in ambiguous grid assignments. In contrast, DETR-based detectors perform global self-attention on coarse feature maps, where tiny objects correspond to only a few low-energy tokens and are thus disadvantaged during bipartite matching. To address these challenges, we propose TinyFormer, a unified YOLO--DETR hybrid real-time detector that combines ViTs, the NMS-free set prediction of DETRs, and the YOLO neck for accurate detection of small objects. TinyFormer introduces a Parallel Bi-fusion Module (PBM) that establishes multiple high-resolution shortcuts from shallow layers to the pyramid neck, preserving fine-grained spatial details during multi-scale fusion. In addition, to mitigate spatial information loss caused by stride-16 tokenization in standard DETR architectures, we propose a Spatial Semantic Adapter (SSA) that extracts high-resolution cues from early network stages and injects them into transformer token embeddings, enhancing tiny-object localization. Extensive experiments on MS COCO demonstrate that TinyFormer outperforms the state-of-the-art (SoTA) detectors, including the latest YOLO-series, as well as the strong baseline DEIMv2. Specifically, TinyFormer-X achieves an impressive 58.4\% AP without the PBM. Integrating PBM pushes the overall performance to 58.5\% AP and maximizes the utility of the supplemented spatial features, yielding a remarkable 1.6\% AP boost in small object AP$_S$. Moreover, when pre-trained on Objects365, TinyFormer-X-PBM reaches 60.2\% AP, surpassing other Objects365-pretrained methods like RF-DETR while needing much fewer parameters and lower compute. These results validate that TinyFormer offers a superior trade-off between accuracy and efficiency. The code is available at \url{https://github.com/mmpmmpmmpjosh/TinyFormer}.

\end{abstract}


\section{Introduction}
\label{sec:intro}


\begin{figure}[h]
    \centering
    \begin{subfigure}{0.32\linewidth}
        \centering
        \includegraphics[width=\linewidth]{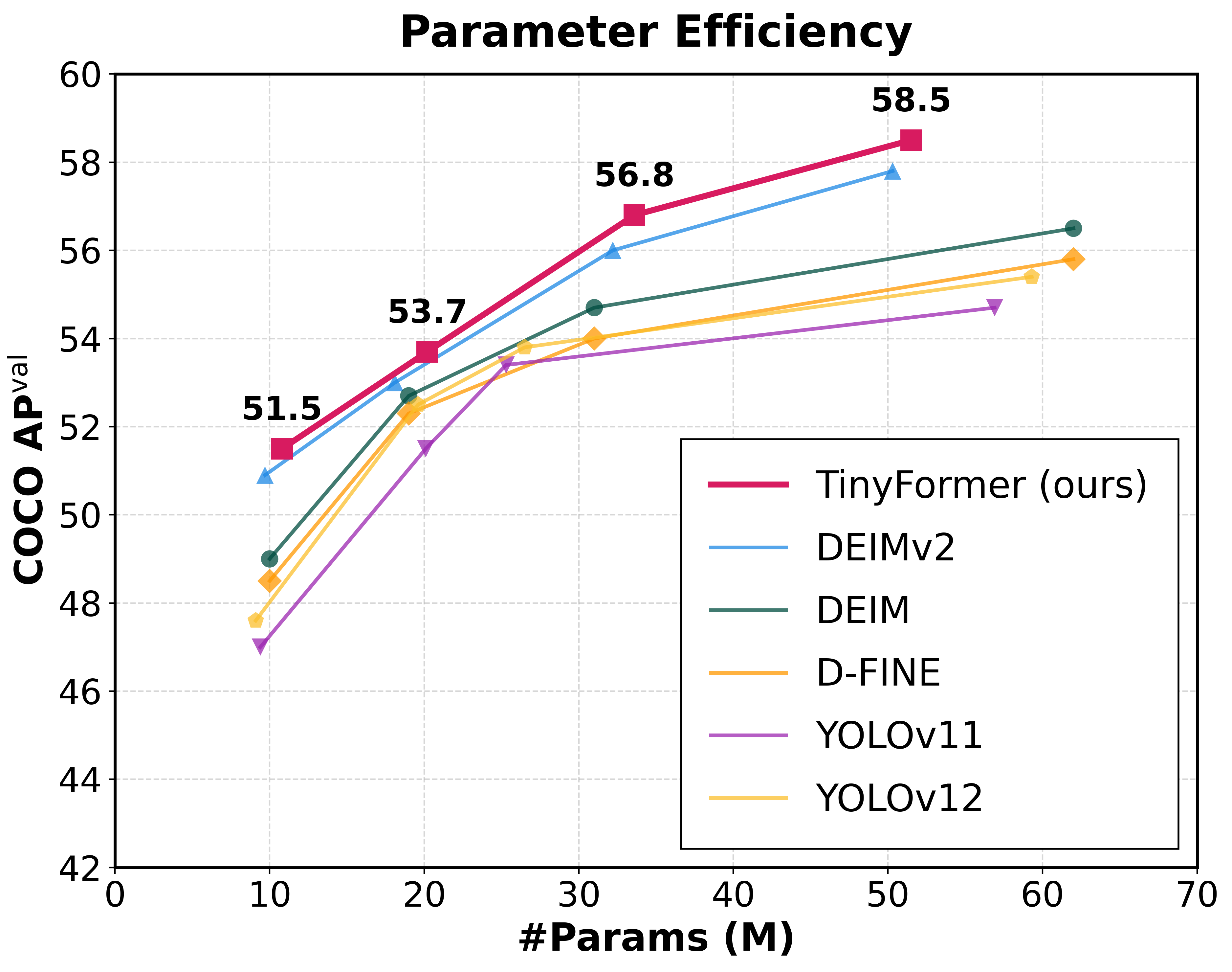}
        \caption{Performance v.s. Params.}
    \end{subfigure}
    \hspace{2mm}
    \begin{subfigure}{0.32\linewidth}
        \centering
        \includegraphics[width=\linewidth]{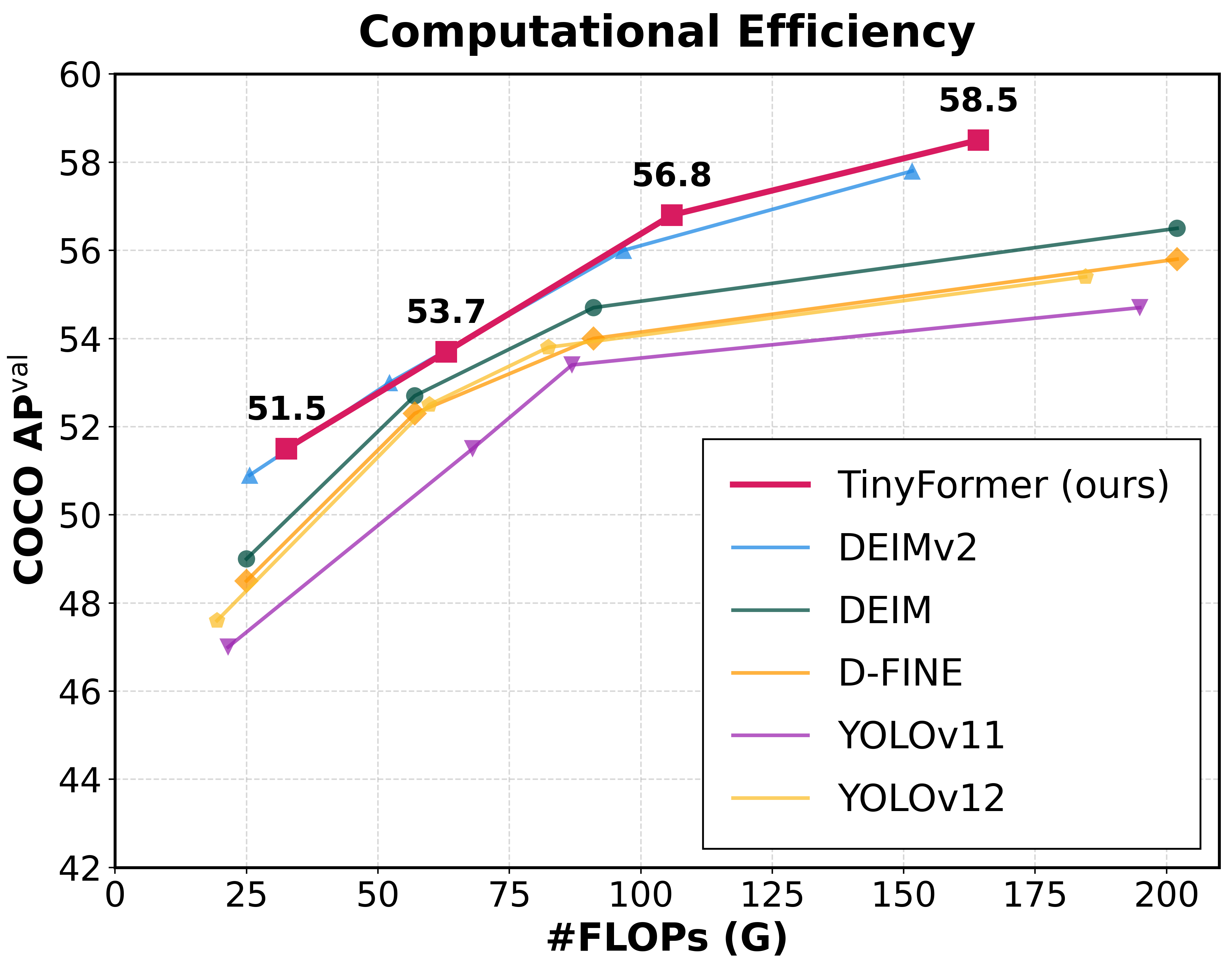}
        \caption{Performance v.s. FLOPs.}
    \end{subfigure}
    \hspace{2mm}
    \begin{subfigure}{0.32\linewidth}
        \centering
        \includegraphics[width=\linewidth]{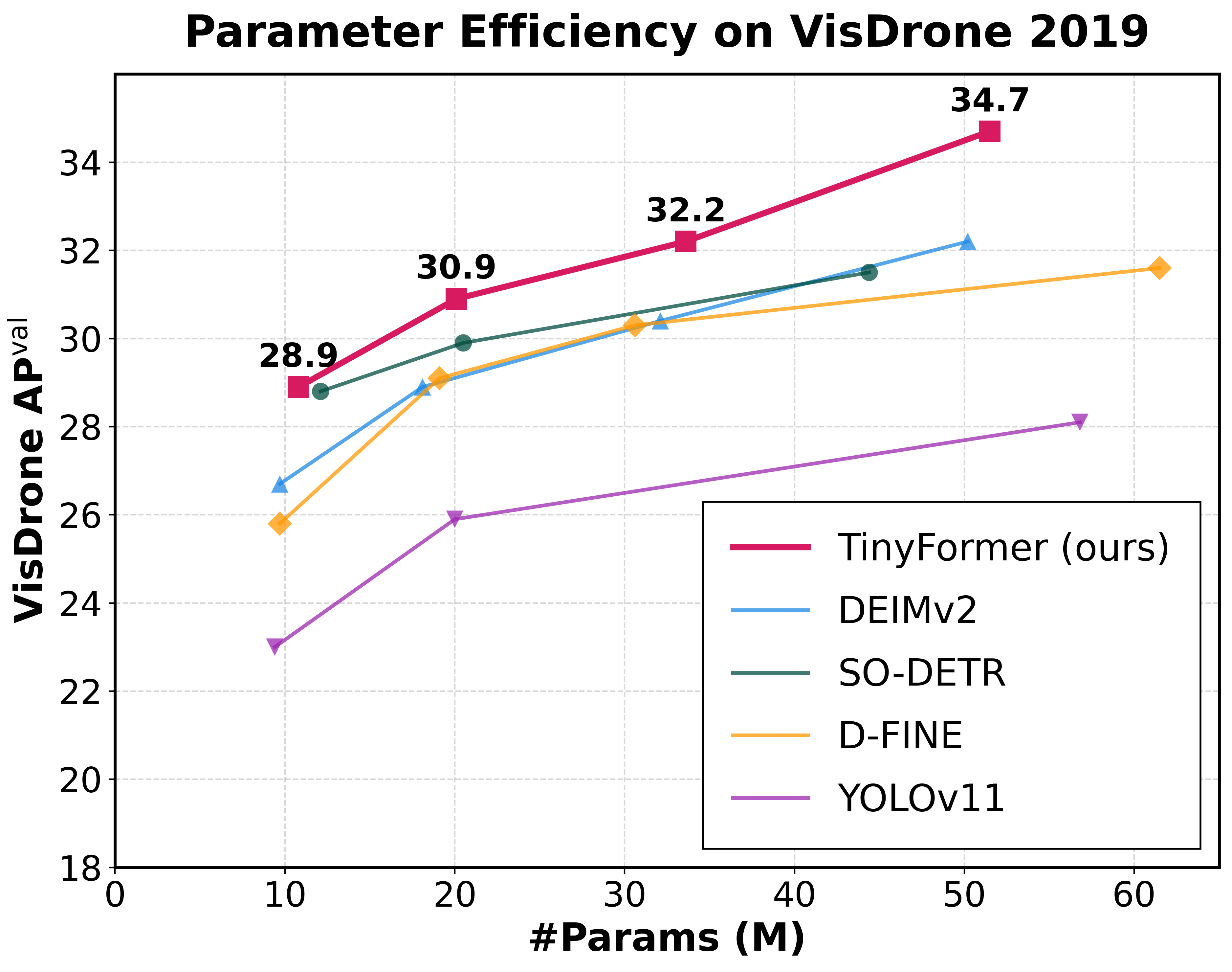}
        \caption{Performance on VisDrone.} 
    \end{subfigure}
    
    \caption{\textbf{Comparison of TinyFormer with the state-of-the-art detectors on COCO~\cite{lin2014microsoft} and VisDrone~\cite{zhu2021detection}.} By effectively recovering spatial priors, TinyFormer explicitly surpasses CNN-based YOLOs in small object detection while maintaining higher overall AP than ViT-based DETRs.}
    \label{fig:front_comp}
\end{figure}

Real-time object detection is often seen as solved, with near-saturated benchmarks. Yet small object detection remains difficult. In safety-critical tasks like autonomous driving~\cite{li2020deep} and surveillance~\cite{wang2020surveiledge}, small objects dominate long-range and crowded scenes, but their accuracy lags behind larger ones.
We argue this stems not from optimization or heuristics, but from structural biases in dominant paradigms. CNN detectors (\eg, YOLO~\cite{yolo11,sapkota2025yolo26,tian2025yolov12}) lose fine spatial detail via aggressive downsampling, while grid assignment and Non-Maximum Suppression (NMS) further hurt performance on small objects. DETRs~\cite{carion2020end,lv2024rtdetrv2improvedbaselinebagoffreebies,lv2023detrs} remove NMS and use global attention, but they introduce the token competition bias in favor of large instances, suppressing small ones.
Despite architectural differences, both CNN and Transformer detectors share a key bottleneck: early compression to stride-32 features. This downsampling irreversibly removes high-frequency spatial cues, causing tiny objects to vanish before high-level reasoning. Once lost, such detail cannot be recovered by deeper layers or better matching. We argue this early information loss is the main barrier to small-object detection and must be addressed by preserving spatial detail from the earliest stages.

To address this limitation, we propose \textbf{TinyFormer}, a YOLO–DETR hybrid combining Vision Transformer (ViT) semantics, DETR’s NMS-free prediction, and a YOLO-style neck with explicit spatial preservation. Since FPNs use heavily downsampled features where small objects occupy few cells, we propose the \textbf{Parallel Bi-fusion Module (PBM)} to add parallel high-resolution paths from shallow layers to the feature pyramid neck, preserving fine-grained spatial details during multi-scale fusion. As DETR tokenizes stride-32 features, small objects have few tokens and weak localization. We thus propose the \textbf{Spatial Semantic Adapter (SSA)} to extract fine-grained spatial features from shallow representations and inject them into transformer tokens, complementing ViT semantics with minimal computational overhead. By preventing early spatial loss in both convolutional and transformer pipelines, TinyFormer preserves fine detail while remaining real-time, achieving SoTA results (Fig.~\ref{fig:front_comp}): 60.2\% AP / 43.0\% AP$_S$ on COCO~\cite{lin2014microsoft}, and 34.7\% AP / 24.7\% AP$_S$ on VisDrone2019~\cite{zhu2021detection}, outperforming recent YOLO~\cite{yolo11,sapkota2025yolo26,tian2025yolov12} and DETR~\cite{peng2024d} detectors.

Our main contributions are summarized as follows:
(1) We propose \textbf{TinyFormer}, a hybrid YOLO-DETR architecture that improves tiny object detection by combining the global contextual modeling of Vision Transformers with an enhanced YOLO-style neck. TinyFormer outperforms all recent YOLO-based~\cite{yolo11,sapkota2025yolo26,tian2025yolov12} and DETR-based detectors~\cite{peng2024d} on both COCO and VisDrone2019 datasets.
(2) We introduce the \textbf{Parallel Bi-fusion Module (PBM)} to break the irreversible spatial degradation of traditional FPNs. By establishing multiple direct high-resolution shortcuts to the fusion neck, the PBM ensures pixel-level localization for tiny targets.
(3) We design the \textbf{Spatial Semantic Adapter (SSA)} to explicitly recover the early-stage spatial details typically lost during the ViT tokenization process, seamlessly fusing high-resolution spatial priors with deep semantic features.
(4) TinyFormer is the first real-time detector to surpass 60\% AP and  40.5\% $AP_S$ on the COCO benchmark without relying on pretraining from external datasets. 

\section{Related Work}
\label{sec:related}

\textbf{Convergences of YOLO and DETR.}
Object detection is dominated by CNN-based YOLO \cite{bochkovskiy2020yolov4, yolov8,jocher2020yolov5} and Transformer-based DETR \cite{carion2020end, lv2023detrs}. YOLO is fast but 32× downsampling loses tiny objects, while grid assignment and NMS further hurt them in crowded scenes. DETR~\cite{carion2020end,lv2024rtdetrv2improvedbaselinebagoffreebies,lv2023detrs} removes NMS and uses global attention, but token competition favors large instances, suppressing small ones. Recently, both paradigms are converging toward a balance of efficiency and end-to-end learning. YOLO26~\cite{sapkota2025yolo26} adds Hungarian matching for NMS-free detection, bridging YOLO’s localization and DETR’s simplicity. TinyFormer follows this trend, combining both to improve real-time detection.


\textbf{Backbone Bottleneck for Tiny Object Detection.} 
Both YOLO and DETR are limited by aggressive backbone downsampling. Modern models use stride-32 features, where each cell covers a large input region. This boosts efficiency and semantics but harms tiny objects, which may shrink to one token or vanish. Thus, YOLO suffers poor grid assignment and localization, while DETR has weak token representation and higher attention competition. This stride-32 bottleneck limits tiny object detection in both, explaining why even YOLO26-X and RT-DETRv2 still struggle—a problem not solved by parameter scaling alone.


\textbf{Vision Transformers and Adapters.}
Vision Transformers (ViTs)~\cite{dosovitskiy2020image}, including DINOv3, struggle with tiny objects. Patch tokenization (\eg, 14×14 or 16×16) lowers resolution, causing tiny objects to collapse into one token or blend with background. Global self-attention also favors large regions, suppressing few-token objects. Thus, ViTs are less effective for precise tiny localization. We address this by proposing the Spatial Semantic Adapter (SSA), a spatial branch alongside the ViT. By extracting and injecting fine-grained details, SSA preserves DINOv3’s semantics while restoring high-resolution cues for reliable tiny object detection.


\textbf{Multi-scale Feature Fusion.}
Feature fusion enables scale-invariant detection; FPN \cite{lin2017feature} and PAN \cite{liu2018path} are standard. However, their top-down upsampling cannot recover spatial detail lost in early downsampling, leaving irreversible degradation—a key issue for real-time detectors. We propose the Parallel Bi-fusion Module (PBM). Rather than reconstructing coarse features, PBM adds direct high-resolution shortcuts, injecting raw spatial details (from SSA) via a parallel path. This preserves tiny object boundaries, improving $AP_S$ and overall $AP$.


\section{Method}

\begin{figure}[t]
    \centering
    \includegraphics[width=\linewidth]{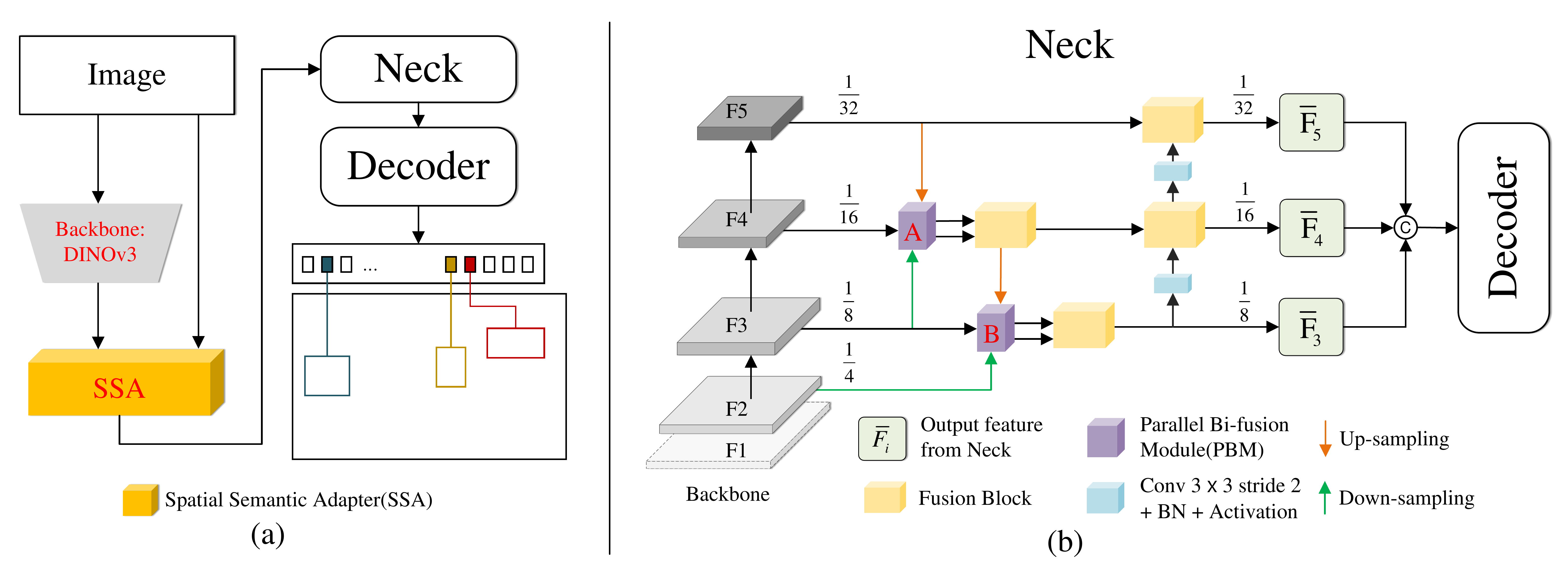}
    \caption{Overall architecture of TinyFormer. The framework consists of a DINOv3~\cite{simeoni2025dinov3} backbone, a multi-scale fusion neck, and an attention-based decoder. (a) Macro-level architecture to highlight the integration of Spatial Semantic Adapter (SSA) and Neck structure. (b) Detailed internal structure of the neck, which processes the enhanced multi-scale features ($F_2, F_3, F_4, F_5$, corresponding to $1/4, 1/8, 1/16$, and $1/32$ scales) provided by SSA. }
    \label{fig:architecture}
\end{figure}
We propose \textbf{TinyFormer}, an end-to-end real-time object detector designed to resolve the failure of current paradigms in localizing tiny objects. As illustrated in Fig.~\ref{fig:architecture}, our pipeline consists of a Vision Transformer backbone, a YOLO-style feature pyramid neck, and an attention-based decoder. TinyFormer introduces two core innovations: the \textbf{Parallel Bi-fusion Module (PBM)}, which establishes bifusion shortcuts in parallel to inject high-resolution spatial details into the neck; \textbf{Spatial Semantic Adapter (SSA)} rescues detailed spatial feature loss caused by the aggressive ViT tokenization process.
Thus, TinyFormer can ensure that tiny targets survive the entire forward pass without being overwhelmed by global context or lost in downsampling.



\begin{figure}[t] 
    \centering
    \includegraphics[width=\linewidth]{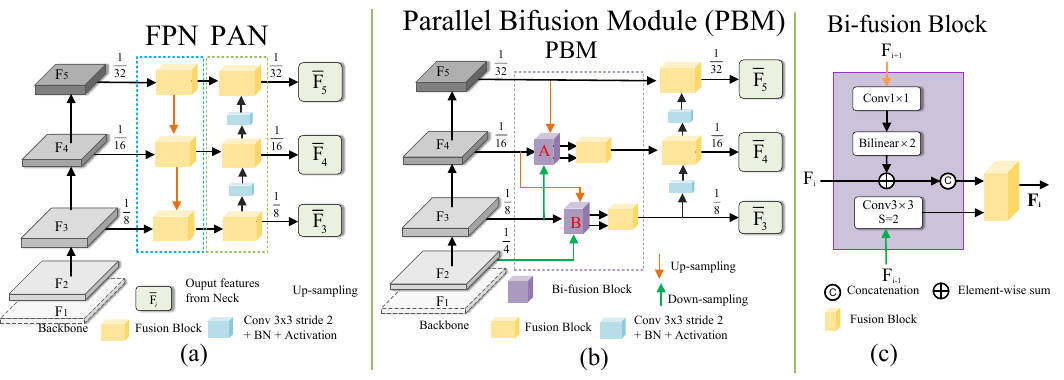} 
    \caption{\textbf{Parallel Bi-fusion Module (PBM).} (a) Standard FPN and PAN structures. (b) PBM (purple) replaces conventional top-down fusion with two parallel bi-fusion blocks that aggregate features from three scales: current ($F_i$), deeper semantic context ($F_{i+1}$), and shallower high-resolution details ($F_{i-1}$), enabling bidirectional feature flow. (c) Architecture of the bi-fusion block.}
    \label{fig:prb_module}
\end{figure}

\begin{figure}[t]
    \centering
    \includegraphics[width=\linewidth]{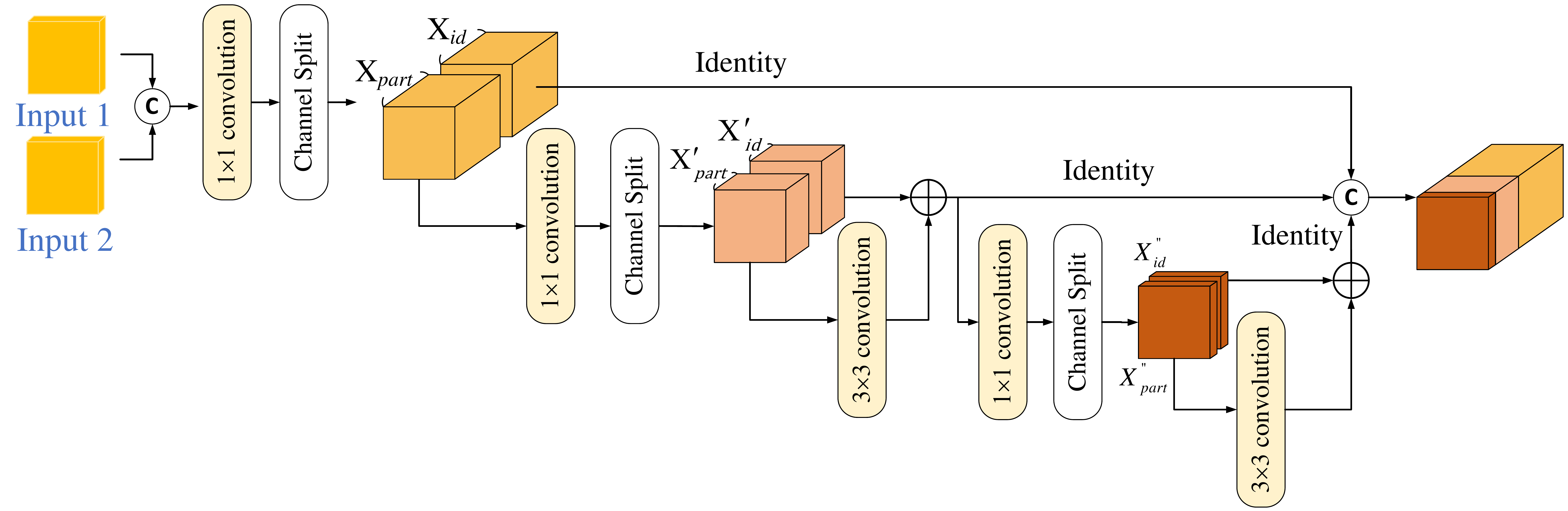}
    \caption{Detailed architecture of the fusion block in Neck.}
    \label{fig:fusion_block}
\end{figure}

\subsection{Parallel Bi-fusion Module (PBM)}
\label{sec:prb}
 
Real-time detectors typically adopt a YOLO-style neck that processes features through a fixed top-down hierarchy ($1/32 \rightarrow 1/16 \rightarrow 1/8$). Although YOLO improves small-object detection via the Path Aggregation Network (PAN) \cite{liu2018path}, which extends the conventional Feature Pyramid Network (FPN) \cite{lin2017feature} with an additional bottom-up pathway for enhanced multi-scale fusion, several limitations remain (Fig.~\ref{fig:prb_module}(a)). First, PAN cannot recover fine-grained spatial information lost during early-stage backbone downsampling. Second, its heavy reliance on top-down upsampling from coarse semantic features often leads to blurred boundaries and degraded texture details. Third, repeated cross-scale feature aggregation may further suppress the already weak responses of tiny objects, particularly in cluttered scenes dominated by large instances.

To address these limitations, we propose the Parallel Bi-fusion Module (PBM), a lightweight bi-fusion connector inspired by PRBNet\cite{9603994}. The PBM is specifically designed to enhance tiny object detection. PBM introduces two bidirectional shortcut pathways that explicitly propagate low-level spatial details from shallow layers (green paths in Fig.~\ref{fig:prb_module}(b)) together with high-level semantic representations from deep layers (orange paths in Fig.~\ref{fig:prb_module}(b)) into the main feature stream, thereby generating fine-grained spatial and semantic features for tiny object detection. In contrast, the neck designs adopted in conventional YOLO-series detectors rely solely on a top-down pathway while discarding bottom-up information flow during cross-scale feature aggregation. As illustrated in Fig.~\ref{fig:prb_module}(b), the conventional top-down cross-scale concatenation is replaced by two parallel bi-fusion blocks (purple blocks). This design enables simultaneous feature interaction across three scales, effectively preserving both semantic context and fine localization cues. Consequently, PBM produces prediction maps with enhanced semantic richness and spatial precision, thereby improving the efficiency and accuracy of detecting both tiny and large objects. The proposed bi-fusion process adopts an ``Align-then-Injection'' mechanism, in which multi-scale features are first spatially aligned and subsequently injected into the target feature stream for effective cross-scale fusion as follows:
\begin{align}
    \tilde{F}_i &= F_{i} + \text{Up}_{\times 2}(\text{Conv}_{1\times1}(F_{i+1})), \label{eq:prbm_align} \\
    \mathbf{F}_{i} &= \mathcal{P} \left( \tilde{F}_i, \text{Conv}_{3\times3, s=2}(F_{i-1}) \right), \label{eq:prbm_supplement}
\end{align}
where $F_{i}$ denotes the features on the target scale, $F_{i+1}$ represents the deep branch that provides high-level semantic context from the heavily downsampled scale, and $\text{Up}_{\times 2}$ denotes spatial upsampling by a factor of 2 (\eg, via bilinear interpolation) to generate the intermediate semantically aligned feature $\tilde{F}_i$. To get
high-resolution spatial details for tiny object detection, we process the shallow spatial branch $F_{i-1}$, which contains raw edge details from the finer scale with a stride-2 convolution. The result and $\tilde{F}_i$ are fed into a fusion block $\mathcal{P}$ to  yield the detail-enriched feature $\mathbf{F}_{i}$. 

The fusion block $\mathcal{P}$ takes two inputs and generates richer representations for object detection. As illustrated in Fig.~\ref{fig:fusion_block}, the two inputs are first concatenated and processed by a $1\times1$ convolution to produce the intermediate feature map $X$. Inspired by CSPNet (Cross Stage Partial Network)~\cite{wang2020cspnet}, the feature map $X$ is then divided into two branches. The partial transformation branch ($X_{\text{part}}$) passes through convolutional layers to extract deeper features, while the identity branch ($X_{\text{id}}$) bypasses these transformations and preserves shortcut information. The two branches are subsequently merged to form the output feature map.
By partially propagating features through the network, this CSP-inspired design reduces computational cost, avoids redundant gradient propagation, and preserves feature diversity. Unlike CSPNet, which partitions channels only once, the proposed fusion block recursively partitions the channels three times. This \textbf{recursive partition strategy} keeps both original and newly generated features at each stage. Finally, multi-level feature maps are fused to produce the output, enabling richer representations and improving the network’s capacity for accurate object detection. 
Another key aspect of PBM is its parallelized bi-fusion design, which allows multi-scale feature aggregation to occur simultaneously rather than sequentially. This parallel structure improves feature representation while maintaining computational efficiency, enabling the detector to capture both fine-grained spatial details and high-level semantic cues for accurate localization across object scales.

\subsection{Spatial Semantic Adapter (SSA)}
\label{sec:ssa}


\begin{figure}[t]
    \centering
    \includegraphics[width=.9\linewidth]{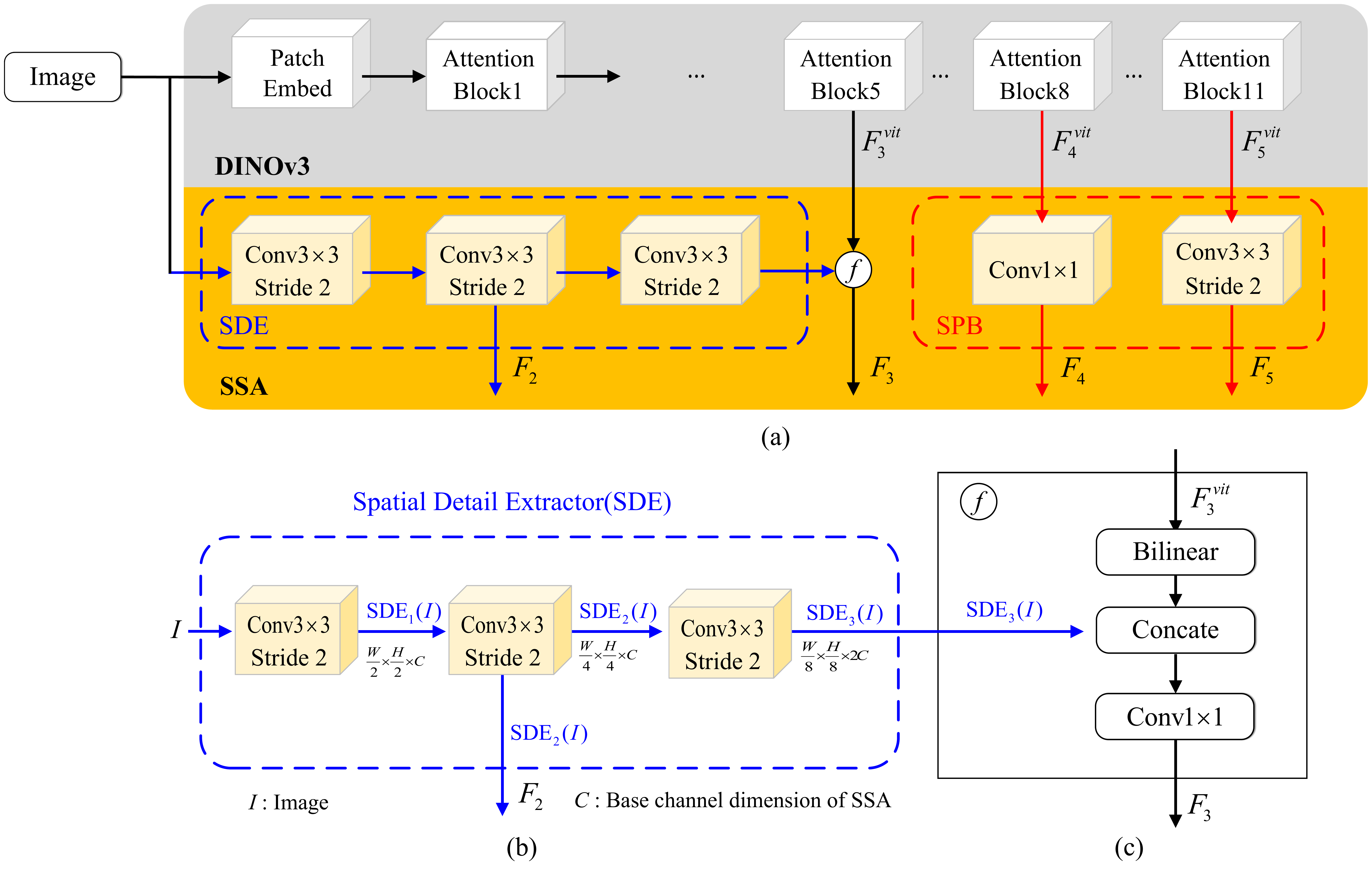}
    \caption{\textbf{Architecture of the Spatial Semantic Adapter (SSA).} (a) SSA integrated with the DINOv3 backbone, featuring the Spatial Detail Extractor (SDE) and Semantic Purification Block (SPB). (b) Detailed structure of the SDE. (c) Intermediate fusion method at the $F_3$ scale. $C$ denotes the base channel dimension.}
    \label{fig:ssa}
\end{figure}

Both CNN- and Transformer-based detectors face challenges when detecting extremely small objects. CNN-based detectors are constrained by the progressive loss of spatial resolution in deeper layers; furthermore, their dependence on Non-Maximum Suppression (NMS) frequently results in missed detections when tiny objects are densely clustered. Transformer-based detectors, in contrast, excel at capturing long-range dependencies to construct robust high-level semantic representations; however, their standard patch embedding, often using large strides, aggressively downsamples the input, causing fine-grained details of tiny objects to be severely degraded or even completely eliminated in early representations.

To address these limitations, we introduce the \textbf{Spatial Semantic Adapter (SSA)}, a lightweight dual-branch module designed to separate spatial detail preservation from semantic representation learning (Fig.~\ref{fig:ssa} (a)). Instead of relying solely on the ViT backbone, SSA includes a \textbf{Spatial Detail Extractor (SDE)} to explicitly preserve high-resolution localization cues and a \textbf{Semantic Purification Block (SPB)} to refine transformer features for robust semantic representation from the deeper layer of pretrained ViT. By integrating these two complementary sources, SSA provides both precise spatial details, thereby improving tiny object detection and mantain reliable semantic context.

The aggressive downsampling during tokenization significantly degrades the discriminative signatures of tiny objects, making it extremely difficult to recover such fine-grained details from deeper feature representations. To address this limitation, SDE operates directly on the raw input image through a sequence of lightweight convolutional layers (as illustrated in Fig.~\ref{fig:ssa}(b)), extracting high-grained spatial priors before substantial information degradation occurs. Formally, let $\text{SDE}_{n}(I)$ denote the feature representation of input image $I$ after passing through $n$ successive $3\times3$ convolutional blocks with stride 2, as shown in Fig.~\ref{fig:ssa}(b):
\begin{equation}
\text{SDE}_{n}(I) = Conv_{3\times3}^{s=2}(\text{SDE}_{n-1}(I)),
\end{equation}
where $\text{SDE}_{0}(I)=I$. This design preserves fine-grained spatial cues at early stages, ensuring that critical details remain accessible to the Parallel Bi-fusion Module (PBM) for effective multi-scale feature fusion and improved tiny object detection. Then, the shallow feature map $F_2$ is defined as
\begin{align}
	F_2 &= \text{SDE}_2(I). 
\end{align}

To effectively combine spatial and semantic information, SSA performs feature fusion at an intermediate level (illustrated in Fig.~\ref{fig:ssa}(a) and (c)):
\begin{align}
	F_3 = \text{Conv}_{1\times1} \left( \text{Concat} \left[ \text{SDE}_3(I), \text{Up}(F^{vit}_3) \right] \right).
\end{align}
This fusion strategy is designed to prevent injecting spatial information too early, as it provides limited benefit due to insufficient semantic context, while introducing it at deeper stages may interfere with already well-formed semantic representations. By fusing at an intermediate scale, SSA achieves a balance between localization precision and semantic richness.

While SDE focuses on preserving spatial fidelity, SPB is designed to address the inconsistency and redundancy of ViT features in intermediate transformer representations when directly used for dense prediction. Raw ViT features are not inherently optimized for detection tasks and may contain noise or misaligned feature distributions before the multiscale fusion in FPN \cite{lin2017feature}. To alleviate this issue, SPB refines the high-level semantic features through lightweight convolutional projections, improving their compatibility with the detection pipeline. The semantic features are formulated as:
\begin{align}
	F_4 &= \text{Conv}_{1\times1}(F^{vit}_4), \\
	F_5 &= \text{Conv}_{3\times3}^{ s=2}(F^{vit}_5),
\end{align}
where $F^{vit}_i$ is the intermediate representations from ViT backbone to construct the $i$-th feature scale.

\begin{table}[t]
\centering
\caption{\textbf{Comparison with real-time object detectors on COCO val2017.} All models are measured on a 3090 GPU with an image size of $640\times640$ and batch size 1. \textbf{Bold} indicates best performance.}
\label{tab:comparison}
\begin{adjustbox}{width=\textwidth}
\renewcommand{\arraystretch}{1.0} 

\begin{tabular}{l | ccc | ccc ccc}
\toprule
\textbf{Method} & \textbf{Params (M)} & \textbf{FLOPs (G)} & \textbf{Lat. (ms)} & \textbf{AP} & \textbf{AP$_{50}$} & \textbf{AP$_{75}$} & \textbf{AP$_S$} & \textbf{AP$_M$} & \textbf{AP$_L$} \\

\midrule
YOLOv10-S~\cite{wang2024yolov10} & 7.2 & 21.6 & 1.41 & 46.3 & 63.0 & 50.4 & 26.8 & 51.0 & 63.8 \\
YOLO11-S~\cite{jocher2024yolov11} & 9.4 & 21.5 & 1.65 & 46.9 & 63.9 & 50.7 & 29.0 & 51.7 & 64.4 \\
YOLOv12-S-turbo~\cite{tian2025yolov12} & 9.2 & 21.4 & 1.90 & 47.5 & 64.1 & - & - & - & - \\
YOLOv13-S~\cite{yolov13} & 9.0 & 20.8 & 2.57 & 48.0 & 65.2 & 52.0 & - & - & - \\
D-FINE-S~\cite{peng2024d} & 10.0 & 25.0 & 1.93 & 48.5 & 65.6 & 52.6 & 29.1 & 52.2 & 65.4 \\
DEIMv2-S~\cite{huang2025deimv2} & 9.7 & 25.6 & 2.34 & 50.9 & 68.3 & 55.1 & 31.4 & \textbf{55.3} & \textbf{70.3} \\

\rowcolor{Gray}
 \textbf{TinyFormer-S (Ours)} & 9.8 & 25.1 & 2.33 & 51.3 & 68.7 & 55.5 & 32.4 & 55.0 & 70.2 \\
 \rowcolor{Gray}
 \textbf{TinyFormer-S-PBM (Ours)} & 10.8 & 32.6 & 2.36 & \textbf{51.5} & \textbf{68.9} & \textbf{55.7} & \textbf{32.6} & \textbf{55.3} & 69.8 \\

\midrule

RT-DETRv2-S~\cite{lv2024rtdetrv2improvedbaselinebagoffreebies} & 20.0 & 60.0 & 1.81 & 48.1 & 65.1 & 52.1 & 30.2 & 51.2 & 64.2 \\
YOLOv10-M~\cite{wang2024yolov10} & 15.4 & 59.1 & 2.24 & 51.1 & 68.1 & 55.8 & 33.8 & 56.5 & 67.0 \\
YOLO11-M~\cite{jocher2024yolov11} & 20.1 & 68.0 & 2.31 & 51.2 & 67.9 & 55.3 & 33.0 & 56.7 & 67.5 \\
D-FINE-M~\cite{peng2024d} & 19.0 & 57.0 & 2.51 & 52.3 & 69.8 & 56.4 & 33.2 & 56.5 & 70.2 \\
YOLOv12-M-turbo~\cite{tian2025yolov12} & 19.6 & 59.8 & 2.66 & 52.6 & 69.5 & - & - & - & - \\
DEIMv2-M~\cite{huang2025deimv2} & 18.1 & 52.2 & 3.09 & 53.0 & 70.2 & 57.5 & 34.2 & 57.4 & \textbf{71.5} \\

 \rowcolor{Gray}
 \textbf{TinyFormer-M (Ours)} & 18.2 & 51.2 & 3.05 & 53.5 & 70.7 & 58.2 & 35.0 & \textbf{57.9} & \textbf{71.5} \\
 \rowcolor{Gray}
 \textbf{TinyFormer-M-PBM (Ours)} & 20.2 & 63.9 & 3.22 & \textbf{53.7} & \textbf{70.9} & \textbf{58.3} & \textbf{35.1} & 57.8 & 71.3 \\

\midrule
RT-DETRv2-M~\cite{lv2024rtdetrv2improvedbaselinebagoffreebies} & 36.4 & 92.0 & 2.28 & 49.9 & 67.5 & 54.1 & 32.0 & 53.2 & 66.5 \\
YOLOv10-L~\cite{wang2024yolov10} & 24.4 & 120.3 & 3.17 & 53.2 & 70.1 & 58.1 & 35.8 & 58.5 & 69.4 \\
YOLO13-L~\cite{yolov13} & 27.6 & 88.4 & 5.33 & 53.4 & 70.9 & 58.1 & - & - & - \\
RT-DETRv2-L~\cite{lv2024rtdetrv2improvedbaselinebagoffreebies} & 42.7 & 136.0 & 3.09 & 53.4 & 71.6 & 57.4 & 36.1 & 57.9 & 70.8 \\
YOLO11-L~\cite{jocher2024yolov11} & 25.3 & 86.9 & 2.83 & 53.4 & 70.1 & 58.2 & 35.6 & 59.1 & 69.2 \\
D-FINE-L~\cite{peng2024d} & 31.0 & 91.0 & 3.26 & 54.0 & 71.6 & 58.4 & 36.5 & 58.0 & 71.9 \\
YOLOv12-L-turbo~\cite{tian2025yolov12} & 26.5 & 82.4 & 4.10 & 54.0 & 70.6 & - & - & - & - \\
YOLOv10-X~\cite{wang2024yolov10} & 29.5 & 160.4 & 3.81 & 54.4 & 71.3 & 59.3 & 37.0 & 59.8 & 70.9 \\
DEIMv2-L~\cite{huang2025deimv2} & 32.2 & 96.7 & 3.59 & 56.0 & 73.5 & 61.1 & 37.5 & 60.8 & 75.2 \\

 \rowcolor{Gray}
 \textbf{TinyFormer-L (Ours)} & 32.3 & 96.3 & 3.54 & 56.5 & 74.0 & 61.5 & 38.1 & \textbf{61.2} & 75.1 \\
 \rowcolor{Gray}
 \textbf{TinyFormer-L-PBM (Ours)} & 33.6 & 105.9 & 3.72 & \textbf{56.8} & \textbf{74.3} & \textbf{61.8} & \textbf{39.0} & \textbf{61.2} & \textbf{75.5} \\

\midrule
RT-DETRv2-X~\cite{lv2024rtdetrv2improvedbaselinebagoffreebies} & 76.0 & 259.0 & 4.42 & 54.3 & 72.8 & 58.8 & 35.8 & 58.8 & 72.1 \\
YOLO11-X~\cite{jocher2024yolov11} & 56.9 & 194.9 & 4.01 & 54.6 & 71.6 & 59.5 & 37.7 & 59.7 & 70.2 \\
YOLO13-X~\cite{yolov13} & 64.0 & 199.2 & 7.39 & 54.8 & 72.0 & 59.8 & - & - & - \\
YOLOv12-X-turbo~\cite{tian2025yolov12} & 59.3 & 184.6 & 5.90 & 55.7 & 72.2 & - & - & - & - \\
D-FINE-X~\cite{peng2024d} & 61.6 & 202.2 & 4.71 & 55.8 & 73.7 & 60.2 & 37.3 & 60.5 & 73.4 \\
DEIMv2-X~\cite{huang2025deimv2} & 50.3 & 151.6 & 4.72 & 57.8 & 75.4 & 63.2 & 39.2 & 62.9 & 75.9 \\
\rowcolor{Gray}
 \textbf{TinyFormer-X (Ours)} & 49.8 & 151.1 & 4.63 & 58.4 & \textbf{75.9} & \textbf{64.1} & 39.3 & 63.1 & 76.5 \\
\rowcolor{Gray}
 \textbf{TinyFormer-X-PBM (Ours)} & 51.5 & 164.2 & 4.81 & \textbf{58.5} & \textbf{75.9} & \textbf{64.1} & \textbf{40.9} & \textbf{63.2} & \textbf{76.6} \\
\midrule
\rowcolor{Gray}
 \textbf{TinyFormer-XL-PBM (Ours)} & 125.5 & 437.9 & 7.91 & 60.6 & 78.0 & 66.3 & 43.4 & 65.9 & 77.9 \\

\bottomrule
\end{tabular}
\end{adjustbox}
\end{table}



\section{Experiment}
\label{sec:experiments}

\begin{table}[t]
\centering
\caption{\textbf{Performance comparison of models pre-trained on Objects365.} All models are fine-tuned on the MS COCO train2017 split and evaluated on the val2017 split. Latency measurements are conducted on a single RTX 3090 GPU with a batch size of 1. \textbf{Bold} indicates best performance.}

\label{tab:objects365}
\begin{adjustbox}{width=\textwidth}
\renewcommand{\arraystretch}{1.15}
\setlength{\tabcolsep}{4pt}
\begin{tabular}{l | c | ccc | ccc ccc}
\toprule
\textbf{Method} & \textbf{Input} & \textbf{Params (M)} & \textbf{FLOPs (G)} & \textbf{Lat. (ms)} & \textbf{AP} & \textbf{AP$_{50}$} & \textbf{AP$_{75}$} & \textbf{AP$_S$} & \textbf{AP$_M$} & \textbf{AP$_L$} \\
\midrule
YOLO26-X~\cite{sapkota2025yolo26} (nms) & $640\times640$ & 55.7 & 193.9 & 4.11 & 57.5 & 75.0 & 62.8 & 41.8 & 62.2 & 73.4 \\
YOLO26-X~\cite{sapkota2025yolo26} (w/o nms) & $640\times640$ & 55.7 & 193.9 & 3.81 & 56.9 & 74.5 & 62.2 & 41.4 & 61.2 & 72.8 \\
RF-DETR-XL~\cite{rf-detr} & $700\times700$ & 126.4 & 299.3 & - & 58.6 & 77.4 & 63.8 & 40.3 & 63.9 & 76.2 \\
D-FINE-X~\cite{peng2024d} & $640\times640$ & 62.0 & 202.0 & 4.71 & 59.3 & 76.8 & 64.6 & 42.3 & 64.2 & 76.4 \\

\rowcolor{Gray}
\textbf{TinyFormer-X-PBM (Ours)} & $640\times640$ & 51.5 & 164.2 & 4.81 & \textbf{60.2} & \textbf{77.6} & \textbf{65.8} & \textbf{43.0} & \textbf{65.1} & \textbf{76.7} \\

\midrule
RF-DETR-2XL~\cite{rf-detr} & $880\times880$ & 126.9 & 438.4 & - & 60.1 & 78.5 & 65.5 & 43.2 & 64.9 & 76.2 \\
\rowcolor{Gray}
\textbf{TinyFormer-XL-PBM (Ours)} & $640\times640$ & 125.5 & 437.9 & 7.91 & \textbf{62.5} & \textbf{79.7} & \textbf{68.4} & \textbf{46.8} & \textbf{67.4} & \textbf{78.2} \\
\bottomrule
\end{tabular}
\end{adjustbox}
\end{table}

\subsection{Comparison with State-of-the-Art (SoTA) Methods}
\label{experiments:sota}

Table~\ref{tab:comparison} compares TinyFormer with recent state-of-the-art real-time detectors (\eg, YOLOv11/12, D-FINE, and DEIMv2) on the COCO dataset. Our framework consistently achieves superior accuracy across all model scales while maintaining competitive latency. At the smallest scale, TinyFormer-S achieves 51.3\% AP, outperforming both DEIMv2-S and YOLOv12-S-turbo. At the large-scale end, TinyFormer-X-PBM establishes a new state of the art with 58.5\% AP, surpassing YOLO11-X and DEIMv2-X while using fewer parameters.
To further explore the scaling capability of the proposed architecture, TinyFormer-XL-PBM achieves 60.6\% AP while preserving real-time inference speed. To the best of our knowledge, without relying on pretraining from external datasets, TinyFormer-XL-PBM is the first real-time detector to surpass 60\% AP on the benchmark. 

\textbf{Breakthrough in Small Object Detection (AP$_S$):} TinyFormer excels in localizing tiny instances by synergizing SSA (which mitigates the ViT spatial bottleneck) and PBM (which optimizes multi-scale aggregation). Integrating PBM yields remarkable AP$_S$ gains across all scales. Notably, TinyFormer-X-PBM achieves an unprecedented 40.9\% AP$_S$; that is, a massive +1.7\% over DEIMv2-X and +1.6\% over our base model. Similarly, the L-size model improves from 38.1\% to 39.0\% AP$_S$ with PBM. These results confirm that PBM effectively resolves conventional neck limitations, fully unleashing SSA's spatial features for fine-grained localization.

\subsection{Scaling up with Objects365 Pre-training}
To explore the architectural upper bound, we pre-train TinyFormer on Objects365 prior to COCO fine-tuning. As shown in Table~\ref{tab:objects365}, TinyFormer-X reaches 60.2\% AP at the standard $640 \times 640$ resolution, outperforming all counterparts at the same scale. Scaling further, our largest variant, TinyFormer-XL, achieves an outstanding 62.5\% AP. Compared to RF-DETR-2XL (60.1\% AP), which relies on $880 \times 880$ input resolution, TinyFormer-XL delivers a +2.4\% AP gain at the standard $640 \times 640$ resolution with comparable parameters. Furthermore, it achieves 46.8\% $AP_S$, surpassing RF-DETR-2XL by +3.6\%, while maintaining real-time inference speed.

\begin{table}[t]
    \centering
    \footnotesize
    \caption{\textbf{Compare with real-time object detectors on the VisDrone 2019 validation set.} All models are evaluated with a standard input size of $640\times640$. \textbf{Bold} means the best performance within each scale. AP$_{75}$, AP$_S$, AP$_M$, and AP$_L$ of TinyFormer and baselines are in the appendix. $^\dagger$: Results are cited from SO-DETR.}
    \label{tab:visdrone}
    \renewcommand{\arraystretch}{1.1}
    \begin{threeparttable}
    \resizebox{.75\linewidth}{!}{
    \begin{tabular}{l | cc | cc}
    \toprule
    \textbf{Method} & \textbf{Params (M)} & \textbf{FLOPs (G)} & \textbf{AP} & \textbf{AP$_{50}$} \\
    \midrule
    
    YOLOv9-S~\cite{wang2024yolov9} \tnote{$\dagger$}& 7.2 & 26.7 & 22.7 & 38.3 \\
    YOLOv8-S~\cite{yolov8} \tnote{$\dagger$}& 11.1 & 28.5 & 22.3 & 37.6 \\
    
    YOLOv11-S~\cite{jocher2024yolov11} \tnote{$\dagger$}& 9.4 & 21.3 & 23.0 & 38.7 \\
    DEIMv2-S~\cite{huang2025deimv2} & 9.7 & 25.3 & 26.7 & 45.0 \\
    D-FINE-S~\cite{peng2024d} & 9.7 & 24.9 & 25.8 & 43.4 \\
    SO-DETR-EFV2 (Distilled)~\cite{zhang2025so} \tnote{$\dagger$}& 12.1 & 33.3 & 28.8 & 47.5  \\
    \rowcolor{Gray}
    \textbf{TinyFormer-S-PBM (Ours)} & 10.8 & 32.6 & \textbf{28.9} & \textbf{47.9} \\
    \midrule
    YOLOv11-M~\cite{jocher2024yolov11} \tnote{$\dagger$}& 20.0 & 67.7 & 25.9 & 43.1 \\
    YOLOv9-M~\cite{wang2024yolov9} \tnote{$\dagger$}& 20.1 & 76.8 & 25.2 & 42.0 \\
    YOLOv8-M~\cite{yolov8} \tnote{$\dagger$}& 25.9 & 78.9 & 24.6 & 40.7 \\
    
    
    DEIMv2-M~\cite{huang2025deimv2} & 18.1 & 51.8 & 28.9 & 48.0 \\
    D-FINE-M~\cite{peng2024d} & 19.1 & 56.1 & 29.1 & 48.3 \\
    SO-DETR-R18~\cite{zhang2025so}\tnote{$\dagger$} & 20.5 & 64.3 & 29.9  & 49.0  \\
    \rowcolor{Gray}
    \textbf{TinyFormer-M-PBM (Ours)} & 20.1 & 63.0 & \textbf{30.9} & \textbf{50.0} \\
    \midrule
    RT-DETRv2-L~\cite{lv2024rtdetrv2improvedbaselinebagoffreebies} & 42.7 & 136.0 & 29.0 & 48.3 \\
    D-FINE-L~\cite{peng2024d} & 30.6 & 90.7 & 30.3 & 50.1 \\
    DEIMv2-L~\cite{huang2025deimv2} & 32.1 & 96.4 & 30.4 & 49.9 \\
    
    \rowcolor{Gray}
    \textbf{TinyFormer-L-PBM (Ours)} & 33.6 & 105.6 & \textbf{32.2} & \textbf{52.1} \\
    \midrule
    YOLOv11-X~\cite{jocher2024yolov11} \tnote{$\dagger$}& 56.8 & 194.5& 28.1 & 45.6 \\
    RT-DETRv2-X~\cite{lv2024rtdetrv2improvedbaselinebagoffreebies} & 76.0 & 259.0 & 29.5 & 49.1 \\
    SO-DETR-R50~\cite{zhang2025so}\tnote{$\dagger$} & 44.4 & 161.4 & 31.5  & 51.5  \\
    D-FINE-X~\cite{peng2024d} & 61.5 & 202.2 & 31.6 & 51.5 \\
    DEIMv2-X~\cite{huang2025deimv2} & 50.2 & 151.3 & 32.2 & 52.3 \\
    
    \rowcolor{Gray}
    \textbf{TinyFormer-X-PBM (Ours)} & 51.5 & 163.9 & \textbf{34.7} & \textbf{55.5} \\
    \bottomrule
    \end{tabular}
    }
    \end{threeparttable}
\end{table}

\begin{figure}[t]
    \centering
    \includegraphics[width=.95\linewidth]{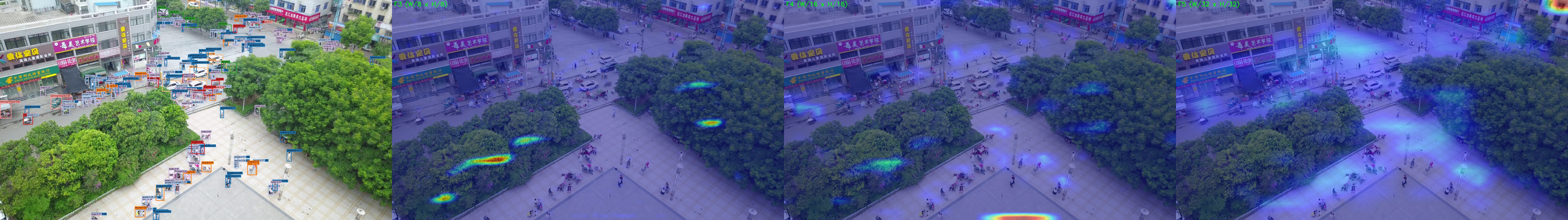} \\
    \vspace{2mm} 
    \includegraphics[width=.95\linewidth]{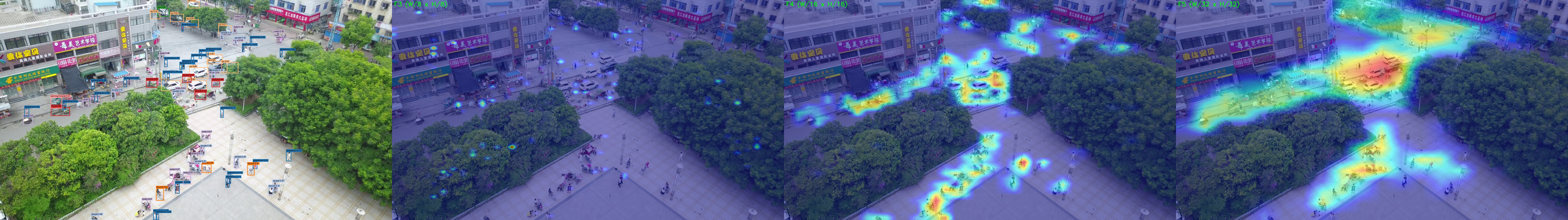}
    \caption{\textbf{Grad-CAM visualization on the VisDrone 2019 val.} Top: baseline DEIMv2-X; bottom: TinyFormer-X-PBM. From left to right, columns show detection results and multi-scale neck activation maps ($\bar{F}_3, \bar{F}_4, \bar{F}_5$, defined in Fig.~\ref{fig:prb_module}, corresponding to $1/8, 1/16, 1/32$ resolutions, respectively).}
    \label{fig:feature_vis}
\end{figure}

\subsection{Generalization on Extreme Scenarios: VisDrone 2019}
\label{experiments:visdrone}

To evaluate TinyFormer's generalization on extreme scenarios dominated by tiny instances, we evaluate its performance on the challenging VisDrone 2019 dataset~\cite{zhu2021detection}. Captured primarily from UAVs, VisDrone features severe scale variations, dense occlusions, and an overwhelming proportion of extremely small objects (often $< 10 \times 10$ pixels).
Table~\ref{tab:visdrone} shows that TinyFormer consistently yields substantial improvements over our strong baseline, DEIMv2, across all scales. Notably, TinyFormer-X-PBM achieves 34.7\% AP and 24.7\% $\text{AP}_S$, outperforming DEIMv2-X by +2.5\% AP and a remarkable +3.2\% $\text{AP}_S$. This exceptional localization capability extends to the M and L variants, which boost $\text{AP}_S$ by +2.4\% and +2.3\% respectively. These robust gains on an aerial-centric benchmark confirm that the synergy between our SSA and PBM modules effectively keeps high-frequency spatial cues. Qualitative visualizations are in Fig.~\ref{fig:feature_vis}

\subsection{Ablation Study}

\textbf{Ablation on Main Components.}
We ablate on TinyFormer's core modules in Table~\ref{tab:ablation_components}. Since PBM needs a 4-scale pyramid, we evaluate its independent performance (Row 3) by creating a surrogate $1/4$ scale input via naive bilinear upsampling from ViT features. We show that while PBM marginally improves AP, the gain in AP$_S$ is limited ($+0.58\%$), exposing a structural bottleneck: aggressive stride-16 tokenization in plain ViTs causes irreversible spatial loss that coarse-feature fusion alone cannot recover. Conversely, isolating SSA within a 3-scale neck (Row 2) greatly boosts AP$_S$ to 39.33\%, confirming that routing raw spatial details from early stages effectively bypasses lossy tokenization. The peak performance (58.50\% AP, 40.94\% AP$_S$) is achieved only when unifying both modules (Row 4). This synergy shows that SSA provides the essential spatial foundation, while PBM ensures its non-diluted distribution across the pyramid to resolve tiny object localization deficiencies.

\begin{table}[t] 
    \centering
    \caption{\textbf{Ablation study on the core components of TinyFormer on MS COCO dataset.} The baseline utilizes 3-scale ViT features via standard bilinear interpolation.}
    \label{tab:ablation_components}
    
    \setlength{\tabcolsep}{1.0mm} 
    \resizebox{.7\linewidth}{!}{
    \begin{tabular}{@{}ccc|cc|c|ccc@{}}
    \toprule
    \textbf{Scales} & \textbf{PBM} & \textbf{SSA} & \textbf{Params (M)} & \textbf{FLOPs (G)} & \textbf{AP} & \textbf{AP$_S$} & \textbf{AP$_M$} & \textbf{AP$_L$} \\
    \midrule
    3 &$\ccross$ &$\ccross$ & 48.8 & 146.8 & 57.13 & 38.50 & 62.67 & 75.62 \\
    3 &$\ccross$ & $\ccheck$ & 49.8 & 151.1 & 58.38 & 39.33 & 63.14 & 76.46 \\
    4 & $\ccheck$ &$\ccross$ & 51.4 & 164.7 & 57.32 & 39.08 & 62.99 & 75.38 \\
    4 & $\ccheck$ & $\ccheck$ & 51.5 & 164.2 & \textbf{58.50} & \textbf{40.94} & \textbf{63.15} & \textbf{76.55} \\
    \bottomrule
    \end{tabular}
    }
\end{table}


\textbf{Generalization of PBM on FPN-based Detectors.} To validate the structural superiority of PBM over traditional top-down FPNs, we plug it into the multi-scale fusion necks of two mainstream SoTA real-time detectors: RT-DETRv2-X and DEIM-X. Table~\ref{tab:gen_prbm} shows that PBM consistently drives up detection performance regardless of the baseline architecture. Notably, the integration of PBM boosts the AP$_S$ of RT-DETRv2-X by +1.62\% along with a +0.20\% gain in overall AP. Similarly, for DEIM-X, it yields a +0.68\% improvement in AP$_S$ and a +0.54\% increase in overall AP. This robust cross-model validation underscores PBM as an effective architectural upgrade for resolving the multi-scale aggregation bottleneck for tiny objects in modern detectors.

\begin{table}[t]
\footnotesize
\centering
\caption{\textbf{Generalization capability of PBM across diverse architectures  on MS COCO dataset.} We integrate the PBM module into the multi-scale fusion necks of representative FPN-based detectors. The consistent performance gains across different backbones (ResNet-101 and HGNetv2) demonstrate that PBM serves as a robust and architecture-agnostic upgrade for real-time detection.}
\label{tab:gen_prbm}
\setlength{\tabcolsep}{3pt} 
\renewcommand{\arraystretch}{1.1}
\resizebox{.77\linewidth}{!}{
\begin{tabular}{l | c | cc | cccc}
\toprule
\textbf{Model} & \textbf{Backbone} & \textbf{Params (M)} & \textbf{FLOPs (G)} & \textbf{AP} & \textbf{AP$_S$} & \textbf{AP$_M$} & \textbf{AP$_L$} \\
\midrule
RT-DETRv2-X~\cite{lv2024rtdetrv2improvedbaselinebagoffreebies} & ResNet-101~\cite{he2016deep} & 76.5 & 260.9 & 53.14 & 32.83 & \textbf{57.74} & 71.70 \\
+ PBM & ResNet-101~\cite{he2016deep} & 79.2 & 287.2 & \textbf{53.34} & \textbf{34.45} & 57.69 & \textbf{72.13} \\
\midrule
DEIM-X~\cite{huang2025deim} & HGNetv2 & 61.6 & 202.6 & 56.02 & 37.50 & 61.10 & \textbf{73.81} \\
+ PBM & HGNetv2 & 62.0 & 214.7 & \textbf{56.56} & \textbf{38.18} & \textbf{61.11} & 73.79 \\
\bottomrule
\end{tabular}
}
\end{table}



\section{Conclusions}
\label{sec:conclusion}

We present TinyFormer, advancing real-time small object detection. Experiments on MS COCO show it overcomes key YOLO and DETR limitations. By integrating the Parallel Bi-fusion Module and the Spatial Semantic Adapter, TinyFormer achieves the SoTA performance: up to 60.2\% AP and 43.0\% AP$_S$ with Objects365 pretraining. TinyFormer surpasses YOLO-series, D-FINE, and DEIMv2 while using much fewer parameters and lower computational costs.

\textbf{Limitations and Future Work.}
TinyFormer is efficient but faces scalability limitations at ultra-high resolutions (\eg, $>1280^2$). Unlike CNNs with linear complexity, Transformer self-attention incurs quadratic complexity with respect to the number of visual tokens, leading to higher computational and memory costs. Future work will explore hardware-aware pruning and linear attention for memory-constrained edge devices.
{
    \small
    \bibliographystyle{plainnat} 
    \bibliography{main}
}

\clearpage
\appendix

\setcounter{table}{0}
\setcounter{figure}{0}
\renewcommand{\thetable}{A\arabic{table}}
\renewcommand{\thefigure}{A\arabic{figure}}

\section*{Appendix}
\label{sec:appendix}



\label{supp:training}
\section{Implementation Details}
\label{experiments:implementation}

We implement TinyFormer in PyTorch 2.5.1 with CUDA 12.2, building upon the DEIMv2~\cite{huang2025deimv2}. To ensure a strictly fair architectural comparison, we adopt the exact same pretrained DINOv3 feature extraction backbone as our baseline DEIMv2. Furthermore, all models are trained following identical hyperparameter schedules and data augmentation pipelines on MS COCO (detailed in the appendix~\ref{sec:coc_train_details}). Our TinyFormer models are evaluated on the MS COCO val2017 dataset with a standard input image size of $640\times640$. For all qualitative visualizations in the main text and appendix, the detection confidence thresholds are set to 0.4 for VisDrone and MS COCO.

For experiments on the VisDrone dataset (Table~\ref{tab:visdrone}), all models trained by us are initialized with their respective COCO pre-trained weights. To ensure a fair evaluation, we train the competing methods (e.g., YOLO-series) strictly following their official COCO training recipes. However, for our proposed TinyFormer and the baseline DEIMv2~\cite{huang2025deimv2}, we explicitly disable the Mosaic augmentation during training to prevent the structural degradation of tiny object characteristics. A detailed ablation study comparing the impact of enabling versus disabling Mosaic augmentation is provided in the appendix.

To ensure consistent speed benchmarking, all latency measurements are conducted on a single NVIDIA RTX 3090 GPU using PyTorch 2.5.1 and CUDA 12.2 with a batch size of 1. Regarding the training phase, all models (with the exception of the XL variant) and the baseline were trained on a single NVIDIA RTX 3090 GPU. The TinyFormer-XL was trained using two NVIDIA H100 GPUs to accommodate its higher computational requirements. We maintained a consistent total batch size of 32 across all models and baseline configurations to ensure a fair comparison. For architectural configurations omitting the Parallel Bi-fusion Module (PBM), the Spatial Semantic Adapter (SSA) explicitly restricts its spatial injection to the $1/8$ fusion level ($F_3$), omitting the $1/4$ scale ($F_2$) pathway to remain compatible with standard 3-scale necks. Detailed training configurations and scaling protocols for the large-scale Objects365~\cite{shao2019objects365} pre-training and fine-tuning are also provided in the appendix~\ref{sec:supp_impl_details}.

\section{COCO Dataset Training Hyperparameters}
\label{sec:coc_train_details}

\begin{table}[h] 
\centering
\caption{Detailed MS COCO training hyperparameters for the TinyFormer variants (XL, X, L, M, S). Settings are mostly adopted from DEIMv2~\cite{huang2025deimv2}, with specific adjustments for the XL variant. \textit{Back.} denotes the backbone. We use \textit{Local Loss} to denote the Fine-Grained Localization (FGL) Loss and the Decoupled Distillation Focal (DDF) Loss introduced in D-FINE~\cite{peng2024d}.}
\label{tab:baseline_hyperparams}
\footnotesize 
\renewcommand{\arraystretch}{1.1} 
\begin{tabular}{l|ccccc}
\toprule
\textbf{Hyperparameter} & \textbf{XL} & \textbf{X} & \textbf{L} & \textbf{M} & \textbf{S} \\
\midrule
Resolution & $640 \times 640$ & $640 \times 640$ & $640 \times 640$ & $640 \times 640$ & $640 \times 640$ \\
Total Epochs & 58 & 58 & 68 & 102 & 132 \\
Local Loss & \checkmark & \checkmark & \checkmark & \checkmark & \checkmark \\
\midrule
Weight Decay & $1.25 \times 10^{-4}$ & $1.25 \times 10^{-4}$ & $1.25 \times 10^{-4}$ & $1.0 \times 10^{-4}$ & $1.0 \times 10^{-4}$ \\
Base LR & $2.5 \times 10^{-4}$ & $5.0 \times 10^{-4}$ & $5.0 \times 10^{-4}$ & $5.0 \times 10^{-4}$ & $5.0 \times 10^{-4}$ \\
Min LR & $1.25 \times 10^{-4}$ & $2.5 \times 10^{-4}$ & $2.5 \times 10^{-4}$ & $2.5 \times 10^{-4}$ & $2.5 \times 10^{-4}$ \\
Back. LR & $5.0 \times 10^{-6}$ & $1.0 \times 10^{-5}$ & $1.25 \times 10^{-5}$ & $2.5 \times 10^{-5}$ & $2.5 \times 10^{-5}$ \\
Back. Min LR & $2.5 \times 10^{-6}$ & $5.0 \times 10^{-6}$ & $6.25 \times 10^{-6}$ & $1.25 \times 10^{-5}$ & $1.25 \times 10^{-5}$ \\
\midrule
\textbf{Augmentation Stages} & \multicolumn{5}{c}{\textbf{Active Epochs}} \\
\midrule
Mosaic ($p=0.5$) & [4, 29] & [4, 29] & [4, 34] & [4, 49] & [4, 64] \\
MixUp ($p=0.5$) & [4, 29] & [4, 29] & [4, 34] & [4, 49] & [4, 64] \\
CopyBlend ($p=0.5$) & [4, 50] & [4, 50] & [4, 60] & [4, 90] & [4, 120] \\
\bottomrule
\end{tabular}
\end{table}

\section{ Object365 Pre-training and Fine-tuning Hyperparameters}
\label{sec:supp_impl_details}

In this section, we provide the comprehensive training protocols for \textbf{TinyFormer-X-PBM} and \textbf{TinyFormer-XL-PBM}. We denote training durations and warmup periods in \textbf{iterations} (equivalent to training batches).

\subsection{Objects365 Pre-training Protocol}
\label{sec:supp_obj365}

To establish a robust feature representation for tiny objects, we first conduct large-scale pre-training on the Objects365 (V2) dataset~\cite{shao2019objects365}. As summarized in Table~\ref{tab:pretrain_specs_combined}, we employ the AdamW optimizer with a constant learning rate to maintain training stability. To prevent over-fitting, \textbf{TinyFormer-XL-PBM} adopts a more conservative learning rate compared to the X variant.

\begin{table}[h] 
\centering
\caption{Hyperparameters for Objects365 pre-training. $p$ denotes augmentation probability.}
\label{tab:pretrain_specs_combined}
\small 
\renewcommand{\arraystretch}{1.2}
\setlength{\tabcolsep}{10pt} 
\begin{tabular}{l | cc}
\toprule
\textbf{Hyperparameter} & \textbf{X-PBM} & \textbf{XL-PBM} \\
\midrule
Input Resolution & $640 \times 640$ & $640 \times 640$ \\
Total Batch Size & 32 & 32 \\
Total Epochs & 20 & 20 \\
Optimizer & \multicolumn{2}{c}{AdamW ($\beta_1=0.9, \beta_2=0.999$)} \\
\midrule
Base LR (incl. SSA) & $5.0 \times 10^{-4}$ & $2.5 \times 10^{-4}$ \\
Backbone LR (DINOv3) & $1.0 \times 10^{-5}$ & $5.0 \times 10^{-6}$ \\
Weight Decay & $1.25 \times 10^{-4}$ & $1.25 \times 10^{-4}$ \\
LR / EMA Warmup (iters) & 2000 / 1000 & 2000 / 1000 \\
LR Scheduler & Constant & Constant \\
EMA Decay & 0.9999 & 0.9999 \\
\midrule
\textbf{Augmentation Stages} & \multicolumn{2}{c}{\textbf{Active Epochs}} \\
\midrule
Random Photo. Distort ($p=0.5$) & [2, 20] & [2, 20] \\
Random Zoom-out / Crop ($p=0.8$)& [2, 20] & [2, 20] \\
Mosaic~\cite{bochkovskiy2020yolov4} / Mixup~\cite{zhang2017mixup} / CopyBlend~\cite{huang2025deimv2} & \multicolumn{2}{c}{Disabled} \\
\bottomrule
\end{tabular}
\end{table}

\begin{table}[h]
\centering
\caption{Hyperparameters for MS COCO fine-tuning for TinyFormer variants.}
\label{tab:coco_specs_combined}
\small
\renewcommand{\arraystretch}{1.2}
\begin{tabular}{l|cc}
\toprule
\textbf{Hyperparameter} & \textbf{X-PBM} & \textbf{XL-PBM} \\
\midrule
Input Resolution & $640 \times 640$ & $640 \times 640$ \\
Total Batch Size & 32 & 32 \\
Total Epochs & 24 & 24 \\
Optimizer & \multicolumn{2}{c}{AdamW ($\beta_1=0.9, \beta_2=0.999$)} \\
\midrule
Base LR (incl. SSA) & $2.5 \times 10^{-4}$ & $1.25 \times 10^{-4}$ \\
Backbone LR (DINOv3) & $5.0 \times 10^{-6}$ & $2.5 \times 10^{-6}$ \\
Weight Decay & $1.25 \times 10^{-4}$ & $1.25 \times 10^{-4}$ \\
LR / EMA Warmup (iters) & 0 / 0 & 0 / 0 \\
LR Scheduler & Constant & Constant \\
EMA Decay & 0.9999 & 0.9999 \\
\midrule
\textbf{Augmentation Stages} & \multicolumn{2}{c}{\textbf{Active Epochs}} \\
\midrule

Mosaic~\cite{bochkovskiy2020yolov4} ($p=1.0$) & [2, 12] & [2, 12] \\
Mixup~\cite{zhang2017mixup} ($p=0.5$) & [2, 12] & [2, 12] \\
CopyBlend~\cite{huang2025deimv2} ($p=0.5$) & [2, 20] & [2, 20] \\
No-Aug Stage (Final) & Last 4 Epochs & Last 4 Epochs \\
\bottomrule
\end{tabular}
\end{table}

\subsection{MS COCO Fine-tuning Protocol}
\label{sec:supp_coco}

After pre-training, the models are fine-tuned on MS COCO~\cite{lin2014microsoft} for 24 epochs. We adopt a multi-stage augmentation strategy~\cite{huang2025deim} where heavy augmentations are disabled in the final 4 epochs. The fine-tuning parameters for both variants are detailed in Table~\ref{tab:coco_specs_combined}.

\label{supp:arch}
\section{Architectural Specifications of TinyFormer Variants}
\label{sec:supp_arch_specs}

In this section, we provide the detailed architectural configurations for all TinyFormer variants. Our scaling strategy is designed to balance the high-level semantic reasoning of ViT backbones with the fine-grained spatial precision provided by the Spatial Semantic Adapter (SSA) and Parallel Bi-fusion Module (PBM).

\subsection{Backbone Scaling and Channel Alignment}
The core of TinyFormer's scalability lies in the selection of the \textbf{DINOv3} backbone series. For variants requiring extreme efficiency (M and S), we adopt distilled compact transformers~\cite{huang2025deimv2} that maintain the semantic priors of larger models. To ensure structural consistency, we define an \textbf{Adapter Base Channel ($C$)} for the SSA, which serves as the primary hidden dimension for spatial detail extraction and interacts with the neck ($d_{Neck}$) and decoder ($d_{Dec}$) as specified in Table~\ref{tab:arch_details}.

\begin{table}[!hbt]
\centering
\caption{\textbf{Architectural details of TinyFormer variants.} $d_{Back.}$ denotes the backbone's output dimension, and $C$ represents the Adapter Base Channel dimension of our SSA.}
\label{tab:arch_details}
\small
\renewcommand{\arraystretch}{1.2}
\begin{tabular}{l|cc|ccc|cc}
\toprule
\multirow{2}{*}{\textbf{Variant}} & \multicolumn{2}{c|}{\textbf{Backbone}} & \multicolumn{3}{c|}{\textbf{Hidden Dimension}} & \multicolumn{2}{c}{\textbf{Layers}} \\
 & \textbf{Model} & \makecell{\textbf{Adapter Base}\\\textbf{Channel ($C$)}} & $d_{Back.}$ & $d_{Neck}$ & $d_{Dec.}$ & \#Back. & \#Dec. \\
\midrule
 \textbf{XL} & ViT-B\cite{simeoni2025dinov3} & 128 & 768 & 384 & 256 & 12 & 6 \\
\textbf{X} & ViT-S+\cite{simeoni2025dinov3} & 64 & 384 & 256 & 256 & 12 & 6 \\
 \textbf{L} & ViT-S\cite{simeoni2025dinov3} & 32 & 384 & 256 & 256 & 12 & 4 \\
\textbf{M} & ViT-T+~\cite{huang2025deimv2} & 16 & 256 & 256 & 256 & 12 & 4 \\
\textbf{S} & ViT-T~\cite{huang2025deimv2} & 16 & 192 & 192 & 192 & 12 & 4 \\
\bottomrule
\end{tabular}
\end{table}

\subsection{Variant Descriptions}
\begin{itemize}
    \item \textbf{Backbone Selection}: While official DINOv3 releases include B, S+, and S scales, we extend TinyFormer to more compact regimes by employing \textbf{ViT-T+} and \textbf{ViT-T}. These models utilize knowledge distillation~\cite{huang2025deimv2} to preserve strong semantic representations within a significantly reduced parameter budget.
    \item \textbf{Adapter Base Channel ($C$)}: This parameter $C$ dictates the base channel dimension of the convolution layer in the SSA. As demonstrated in our SSA architectural diagrams (see Sec. 3, Fig.~\ref{fig:ssa}) and ablation studies (see Sec. 4, Fig.~\ref{fig:ssa_exp}).
    \item \textbf{Multi-scale Configuration}: The integration of our \textbf{Parallel Bi-fusion Module (PBM)} neck naturally enables a \textbf{4-scale} feature pyramid (incorporating the high-resolution $F_2$ spatial prior). This configuration is primarily utilized in our flagship \textbf{XL} variant to maximize tiny object recall, whereas other variants utilize a 3-scale neck optimized for inference efficiency.
    \item \textbf{d$_{Neck}$}: We define the dimension of the PBM as $d_{Neck}$ to emphasize its role in synergizing spatial and semantic features before the attention-based decoder.
\end{itemize}
\clearpage
\section{Training Configurations on VisDrone 2019}
\label{appendix:visdrone_config} 
To ensure a strictly fair comparison on the VisDrone 2019 dataset, we standardize the training configuration across our proposed TinyFormer and the baseline DEIMv2. Both models are initialized from COCO pre-trained weights to accelerate convergence and strictly follow the default COCO training hyperparameters (detailed in Table~\ref{tab:baseline_hyperparams}), with only two specific exceptions: the deactivation of Mosaic augmentation and a standardized warmup schedule. Notably, for aerial-centric datasets dominated by extremely small instances, aggressive spatial augmentations can severely distort object scales. As demonstrated in Table~\ref{tab:visdrone_mosaic}, disabling Mosaic augmentation consistently yields superior performance; for instance, TinyFormer-X-PBM achieves an AP of 34.7\% without Mosaic compared to 34.4\% when enabled. Consequently, all VisDrone results reported in our main text (Table~\ref{tab:visdrone}) are obtained with Mosaic explicitly disabled. Furthermore, for all baseline comparisons utilizing a learning rate warmup, the duration is standardized to approximately half of an epoch (specifically set to 100 iterations in our implementation).

\begin{table}[h]
\centering
\caption{\textbf{Comparison with the baseline real-time object detector on the VisDrone 2019-val set.} All models are evaluated with a standard input size of $640\times640$. \textbf{Bold} indicates best performance within each scale.}
\label{tab:visdrone_detail}
\renewcommand{\arraystretch}{1.0}
\resizebox{\linewidth}{!}{
\begin{tabular}{l | cc | ccc ccc}
\toprule
\textbf{Method} & \textbf{Param.(M)} & \textbf{FLOPs(G)} & \textbf{AP} & \textbf{AP$_{50}$} & \textbf{AP$_{75}$} & \textbf{AP$_S$} & \textbf{AP$_M$} & \textbf{AP$_L$} \\
\midrule
DEIMv2-S~\cite{huang2025deimv2} & 9.7 & 25.3 & 26.7 & 45.0 & 26.5 & 16.7 & 38.9 & 59.5 \\
\rowcolor{Gray}
\textbf{TinyFormer-S-PBM (Ours)} & 10.8 & 32.6 & \textbf{28.9} & \textbf{47.9} & \textbf{29.0} & \textbf{18.9} & \textbf{40.5} & \textbf{61.4} \\
\midrule
DEIMv2-M~\cite{huang2025deimv2} & 18.1 & 51.8 & 28.9 & 48.0 & 29.1 & 18.5 & 41.7 & 64.0 \\
\rowcolor{Gray}
\textbf{TinyFormer-M-PBM (Ours)} & 20.1 & 63.0 & \textbf{30.9} & \textbf{50.0} & \textbf{31.4} & \textbf{20.9} & \textbf{43.2} & \textbf{67.4} \\
\midrule
DEIMv2-L~\cite{huang2025deimv2} & 32.1 & 96.4 & 30.4 & 49.9 & 31.1 & 19.8 & 43.8 & 64.2 \\
\rowcolor{Gray}
\textbf{TinyFormer-L-PBM (Ours)} & 33.6 & 105.6 & \textbf{32.2} & \textbf{52.1} & \textbf{32.7} & \textbf{22.1} & \textbf{44.7} & \textbf{68.5} \\
\midrule
DEIMv2-X~\cite{huang2025deimv2} & 50.2 & 151.3 & 32.2 & 52.3 & 32.9 & 21.5 & 45.6 & \textbf{67.0} \\
\rowcolor{Gray}
\textbf{TinyFormer-X-PBM (Ours)} & 51.5 & 163.9 & \textbf{34.7} & \textbf{55.5} & \textbf{35.8} & \textbf{24.7} & \textbf{47.5} & 66.1 \\
\bottomrule
\end{tabular}
} 
\end{table}

\begin{table}[h]
\centering
\caption{\textbf{Comparison of TinyFormer trained with or without Mosaic on the VisDrone 2019-val set.} All models are evaluated with a standard input size of $640\times640$. \textbf{Bold} indicates best performance within each scale.}
\label{tab:visdrone_mosaic} 
\renewcommand{\arraystretch}{1.0}
\resizebox{\linewidth}{!}{
\begin{tabular}{l | cc | ccc ccc}
\toprule
\textbf{Method} & \textbf{Param.(M)} & \textbf{FLOPs(G)} & \textbf{AP} & \textbf{AP$_{50}$} & \textbf{AP$_{75}$} & \textbf{AP$_S$} & \textbf{AP$_M$} & \textbf{AP$_L$} \\
\midrule
TinyFormer-L-PBM w/ mosaic & 33.6 & 105.6 & 31.4 & 51.2 & 31.9 & 21.8 & 43.3 & 67.9 \\
\rowcolor{Gray}
\textbf{TinyFormer-L-PBM w/o mosaic} & 33.6 & 105.6 & \textbf{32.2} & \textbf{52.1} & \textbf{32.7} & \textbf{22.1} & \textbf{44.7} & \textbf{68.5} \\
\midrule
TinyFormer-X-PBM w/ mosaic & 51.5 & 163.9 & 34.4 & 55.2 & 35.4 & 24.5 & 46.8 & 65.0 \\
\rowcolor{Gray}
\textbf{TinyFormer-X-PBM w/o mosaic} & 51.5 & 163.9 & \textbf{34.7} & \textbf{55.5} & \textbf{35.8} & \textbf{24.7} & \textbf{47.5} & \textbf{66.1} \\
\bottomrule
\end{tabular}
} 
\end{table}
 \clearpage
\section{Additional Qualitative Comparisons}
\label{appendix:qualitative}

This section evaluates TinyFormer’s feature representation quality across various datasets and object scales through detection results and Grad-CAM visualizations. For all visualizations, the detection confidence threshold is set to 0.4 to avoid visual clutter.

\subsection{Qualitative Detection Results on VisDrone 2019 val}

\begin{figure*}[h]
    \centering
    
    \begin{subfigure}{0.48\linewidth}
        \centering
        \includegraphics[width=\linewidth]{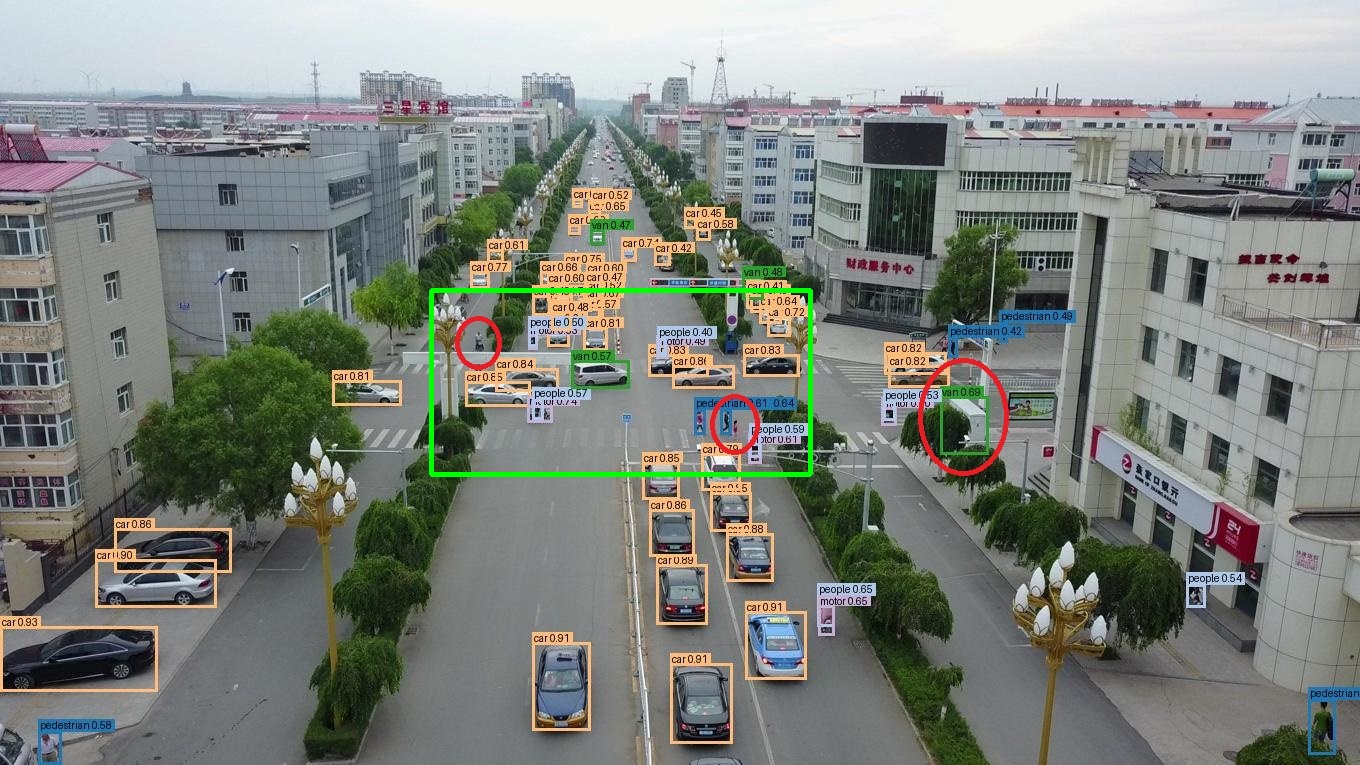}
        \caption{DEIMv2}
        \label{fig:det_1a}
    \end{subfigure}
    \hfill
    \begin{subfigure}{0.48\linewidth}
        \centering
        \includegraphics[width=\linewidth]{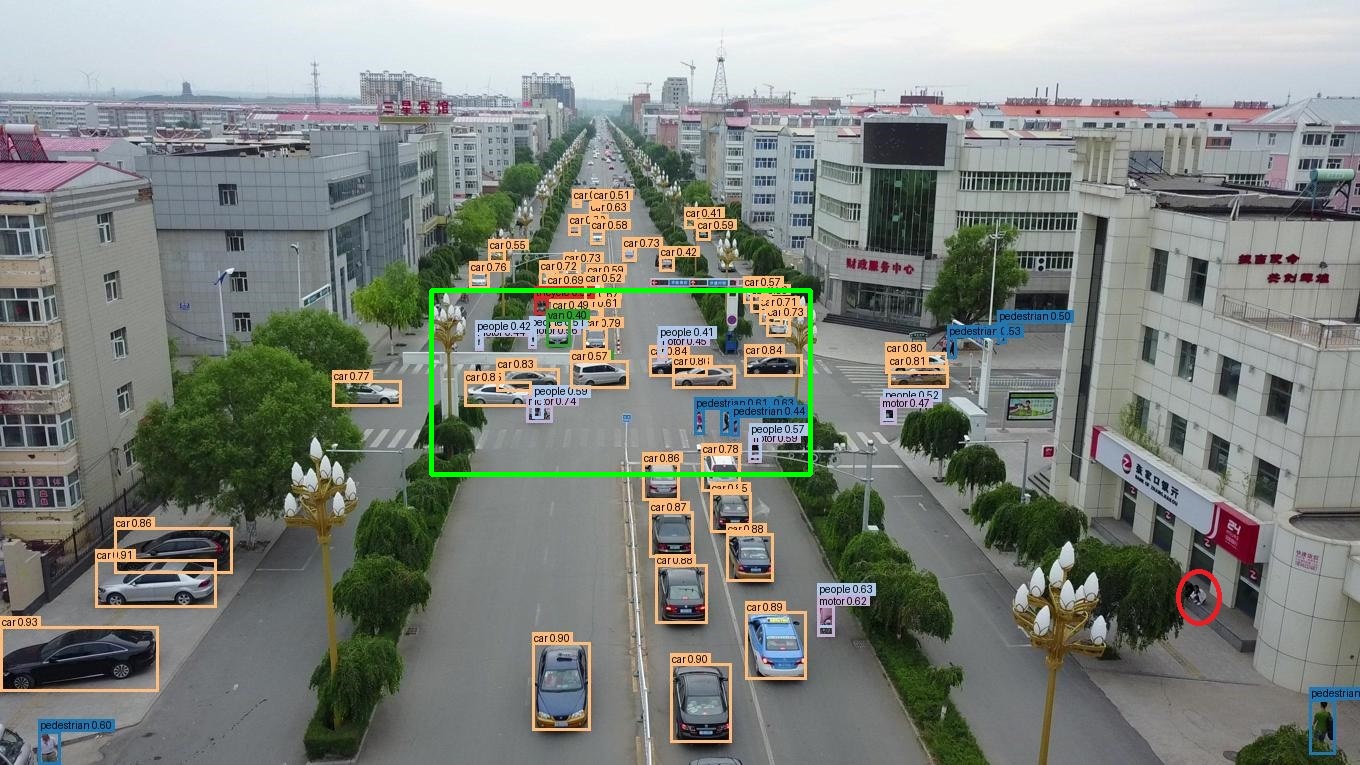}
        \caption{Ours}
        \label{fig:det_1b}
    \end{subfigure}

    \vspace{2mm} 

    \begin{subfigure}{0.48\linewidth}
        \centering
        \includegraphics[width=\linewidth]{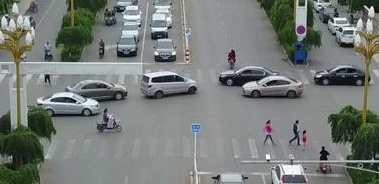}
        \caption{Input Image}
        \label{fig:det_1c}
    \end{subfigure}
    \hfill
    \begin{subfigure}{0.48\linewidth}
        \centering
        \includegraphics[width=\linewidth]{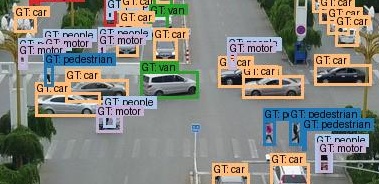}
        \caption{Ground Truth}
        \label{fig:det_1d}
    \end{subfigure}

    \vspace{2mm}

    \begin{subfigure}{0.48\linewidth}
        \centering
        \includegraphics[width=\linewidth]{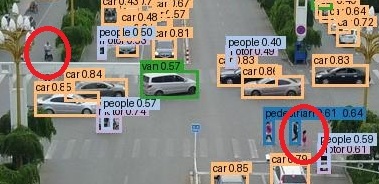}
        \caption{Zoomed-in View of (a)}
        \label{fig:det_1e}
    \end{subfigure}
    \hfill
    \begin{subfigure}{0.48\linewidth}
        \centering
        \includegraphics[width=\linewidth]{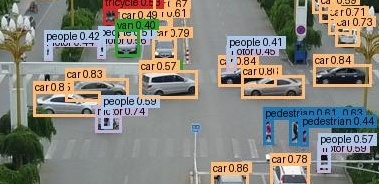}
        \caption{Zoomed-in View of (b)}
        \label{fig:det_1f}
    \end{subfigure}

    \caption{\textbf{Qualitative object detection results.} (a) and (b) show the detection performance of DEIMv2-X and TinyFormer-X-PBM, respectively. (c) Input image and (d) Ground Truth. The red circles in (e) indicate missed detections by the baseline, which are successfully localized by TinyFormer in (f), highlighting our superior recall for tiny objects.}
    \label{fig:supp_qualitative_det1}
\end{figure*}

\begin{figure*}[t]
    \centering
    
    \begin{subfigure}{0.48\linewidth}
        \centering
        \includegraphics[width=\linewidth]{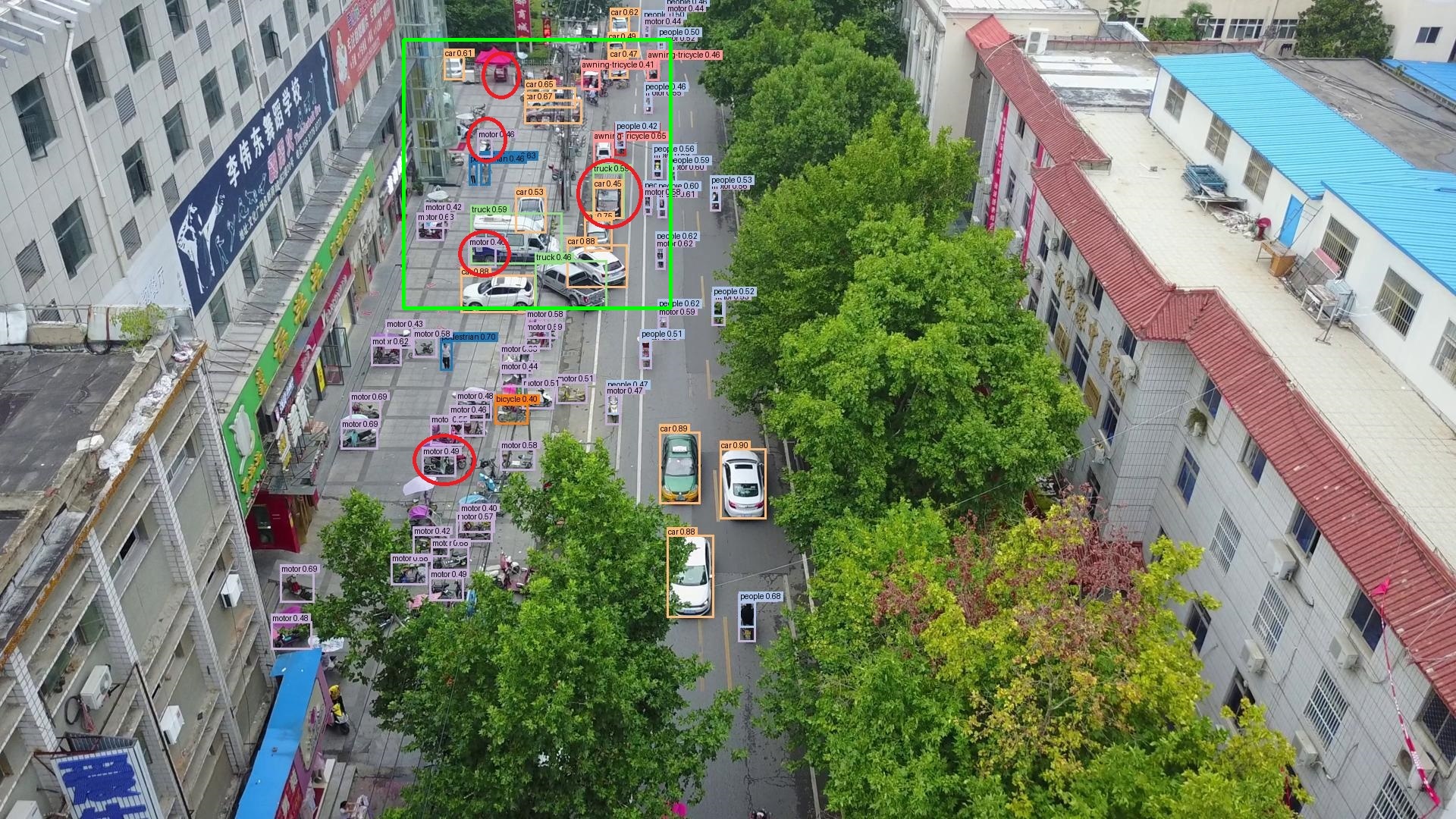}
        \caption{DEIMv2}
        \label{fig:det_2a}
    \end{subfigure}
    \hfill
    \begin{subfigure}{0.48\linewidth}
        \centering
        \includegraphics[width=\linewidth]{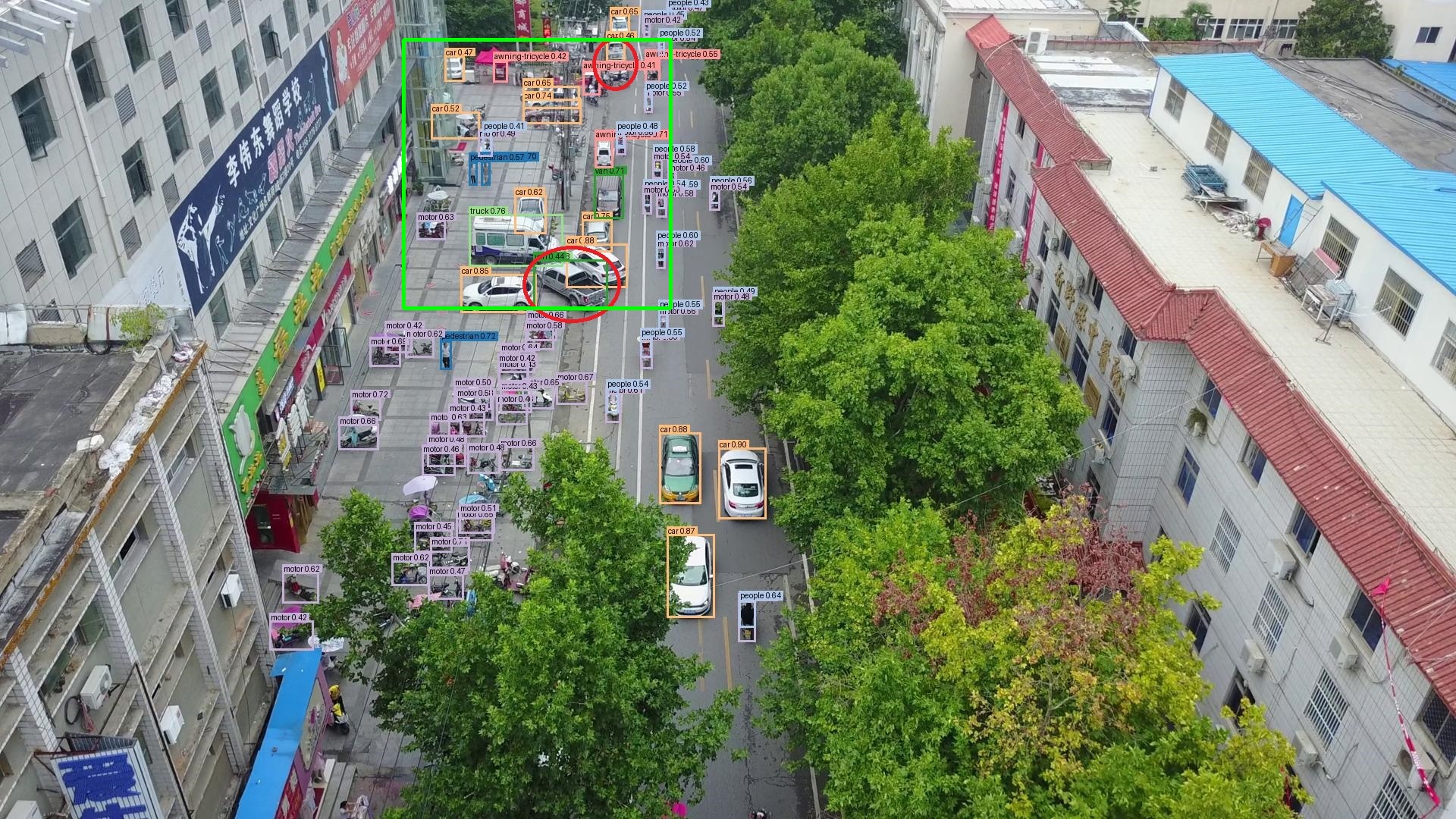}
        \caption{Ours}
        \label{fig:det_2b}
    \end{subfigure}

    \vspace{1.5mm}

    \begin{subfigure}{0.35\linewidth}
        \centering
        \includegraphics[width=\linewidth]{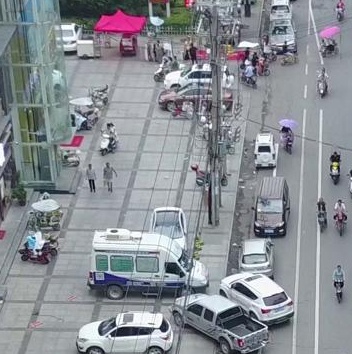}
        \caption{Input Image}
        \label{fig:det_2c}
    \end{subfigure}
    \hspace{0.16\linewidth} 
    \begin{subfigure}{0.35\linewidth}
        \centering
        \includegraphics[width=\linewidth]{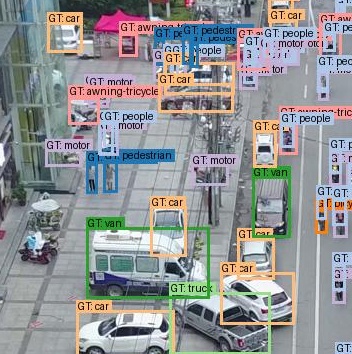}
        \caption{Ground Truth}
        \label{fig:det_2d}
    \end{subfigure}

    \vspace{1.5mm}

    \begin{subfigure}{0.35\linewidth}
        \centering
        \includegraphics[width=\linewidth]{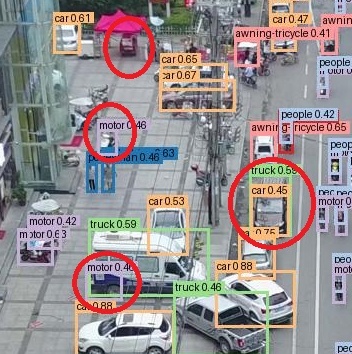}
        \caption{Zoomed-in View of (a)}
        \label{fig:det_2e}
    \end{subfigure}
    \hspace{0.16\linewidth} 
    \begin{subfigure}{0.35\linewidth}
        \centering
        \includegraphics[width=\linewidth]{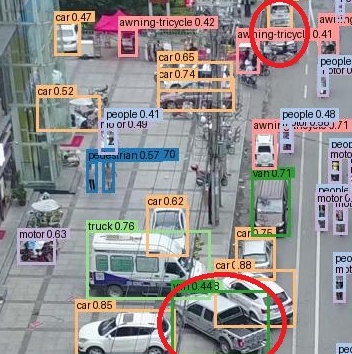}
        \caption{Zoomed-in View of (b)}
        \label{fig:det_2f}
    \end{subfigure}

    \caption{\textbf{Qualitative object detection results.} (a) and (b) show the detection performance of DEIMv2-X and TinyFormer-X-PBM, respectively. (c) Input image and (d) Ground Truth. (e) and (f) provide magnified views of the regions enclosed by green boxes. While the baseline produces redundant and conflicting predictions (incorrectly labeling a pedestrian as both "motor" and "pedestrian"), TinyFormer maintains accurate classification for tiny targets.}
    \label{fig:supp_qualitative_det2}
\end{figure*}

\begin{figure*}[t]
    \centering
    
    \begin{subfigure}{0.48\linewidth}
        \centering
        \includegraphics[width=\linewidth]{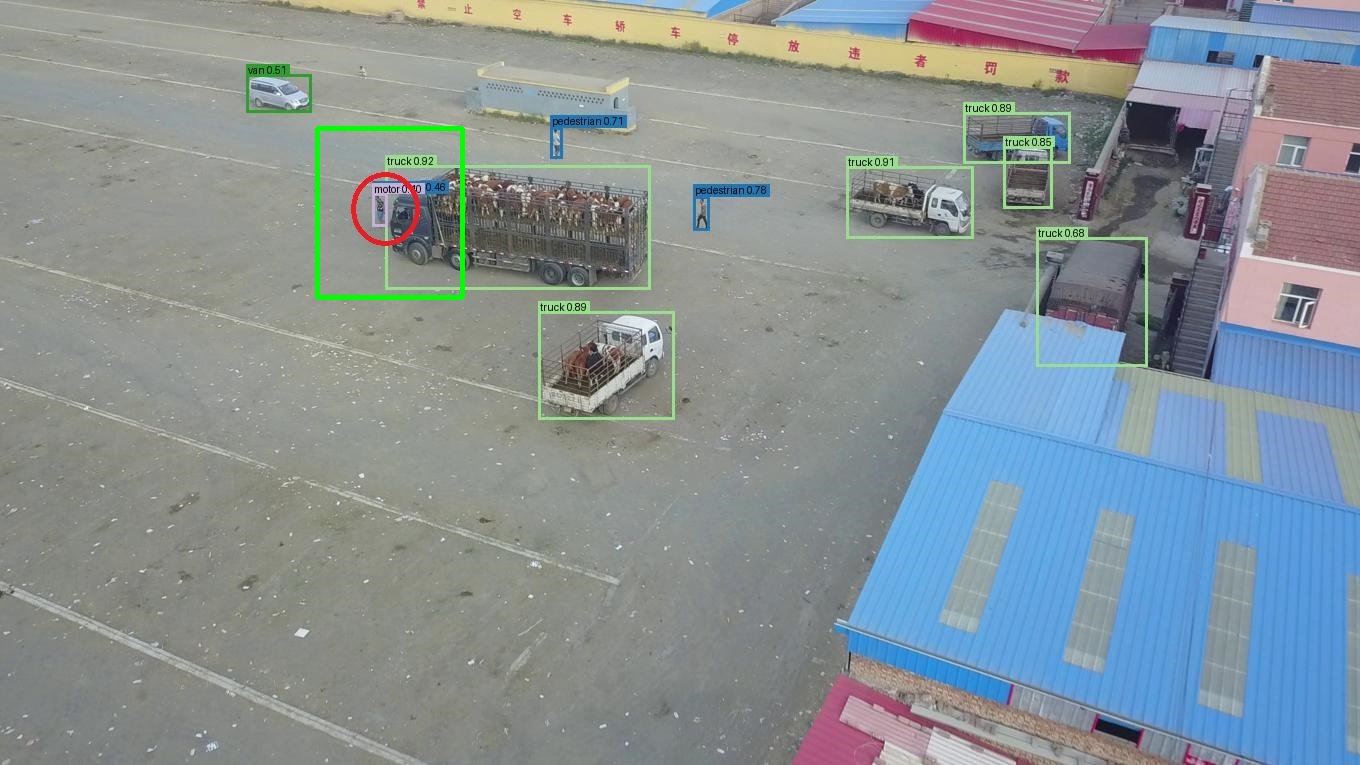}
        \caption{DEIMv2}
        \label{fig:det_3a}
    \end{subfigure}
    \hfill
    \begin{subfigure}{0.48\linewidth}
        \centering
        \includegraphics[width=\linewidth]{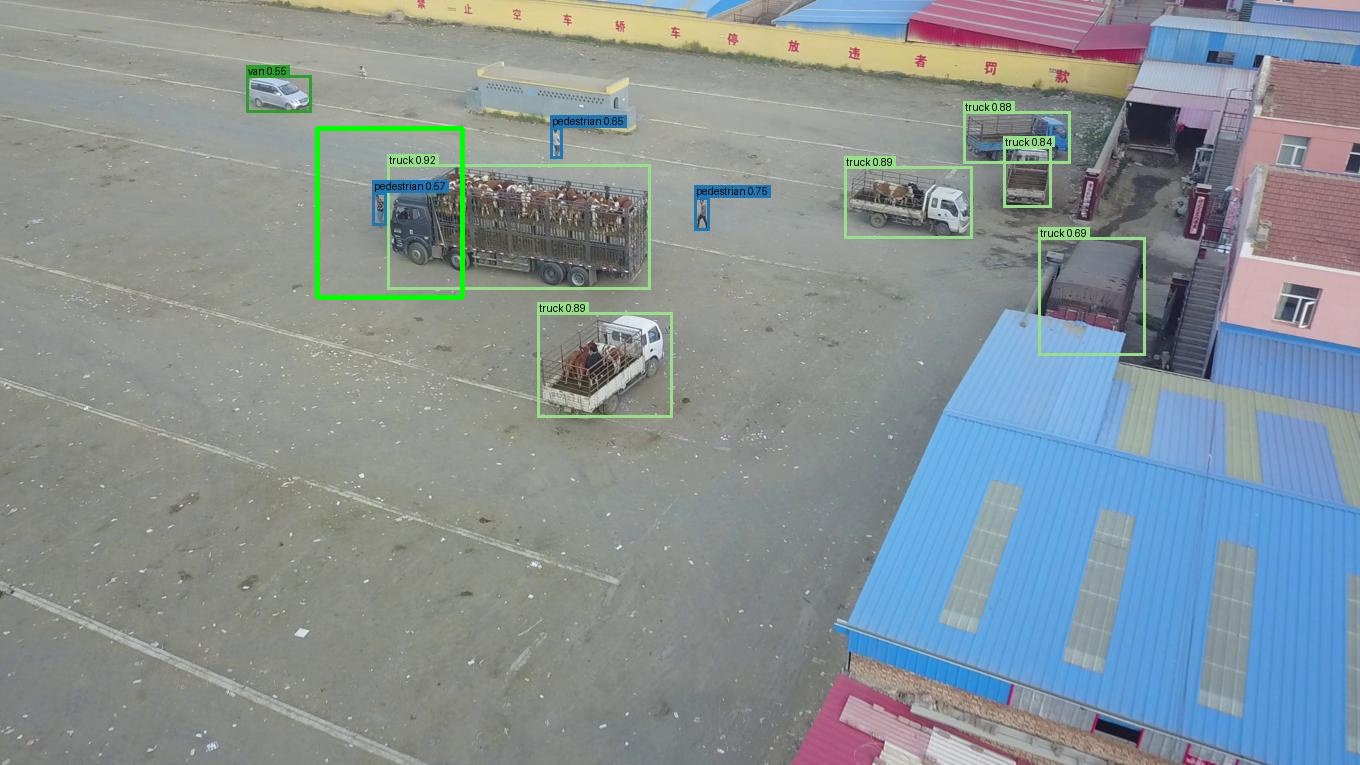}
        \caption{Ours}
        \label{fig:det_3b}
    \end{subfigure}

    \vspace{1.5mm}

    \begin{subfigure}{0.35\linewidth}
        \centering
        \includegraphics[width=\linewidth]{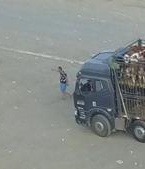}
        \caption{Input Image}
        \label{fig:det_3c}
    \end{subfigure}
    \hspace{0.16\linewidth} 
    \begin{subfigure}{0.35\linewidth}
        \centering
        \includegraphics[width=\linewidth]{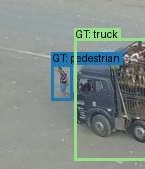}
        \caption{Ground Truth}
        \label{fig:det_3d}
    \end{subfigure}

    \vspace{1.5mm}

    \begin{subfigure}{0.35\linewidth}
        \centering
        \includegraphics[width=\linewidth]{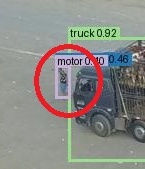}
        \caption{Zoomed-in View of (a)}
        \label{fig:det_3e}
    \end{subfigure}
    \hspace{0.16\linewidth} 
    \begin{subfigure}{0.35\linewidth}
        \centering
        \includegraphics[width=\linewidth]{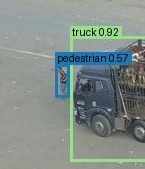}
        \caption{Zoomed-in View of (b)}
        \label{fig:det_3f}
    \end{subfigure}

    \caption{\textbf{Qualitative object detection results.} (a) and (b) show the detection performance of DEIMv2-X and TinyFormer-X-PBM, respectively. (c) Input image and (d) Ground Truth. (e) and (f) provide magnified views of the regions enclosed by green boxes. While the baseline produces redundant and conflicting predictions (incorrectly labeling a pedestrian as both "motor" and "pedestrian"), TinyFormer maintains accurate classification for tiny targets.}
    \label{fig:supp_qualitative_det3}
\end{figure*}

To visually demonstrate the superiority of TinyFormer in capturing extremely small instances, we provide a direct qualitative comparison of object detection results against the baseline DEIMv2 detector. As illustrated in Figs.~\ref{fig:supp_qualitative_det1}--\ref{fig:supp_qualitative_det3}, we select challenging scenes from the VisDrone 2019-val set characterized by dense vehicle distributions and severe scale variations. TinyFormer exhibits significantly higher recall and localization accuracy, particularly in heavily cluttered aerial environments. \textbf{Note that} the red circles focus exclusively on pronounced performance discrepancies in object presence and localization; minor, shared classification ambiguities (\eg, confusing a car with a van) are intentionally left unhighlighted to preserve visual clarity.

\clearpage

\subsection{Grad-CAM Visualizations on VisDrone 2019}
\label{subsec:visdrone_cam}
Following the configuration in the main text, these visualizations are generated across the multi-scale neck outputs ($\bar{F}_3, \bar{F}_4, \bar{F}_5$), as defined in Fig.~\ref{fig:prb_module}, corresponding to $1/8, 1/16$, and $1/32$ spatial resolutions, respectively. From left to right, the columns in each set display the detection results and the corresponding activation maps from the neck outputs. 


\begin{figure}[h]
    \centering
    \begin{subfigure}{\linewidth}
        \centering
        \includegraphics[width=.95\linewidth]{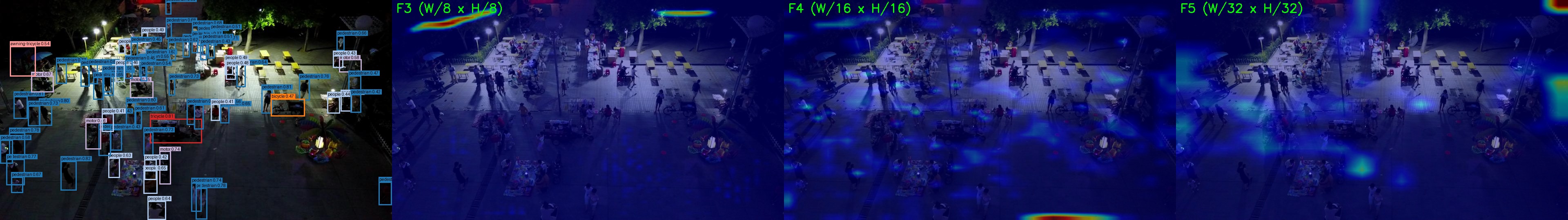}
        \caption{DEIMv2}
        \label{fig:cam_set1_a}
    \end{subfigure} \\
    \vspace{2mm} 
    \begin{subfigure}{\linewidth}
        \centering
        \includegraphics[width=.95\linewidth]{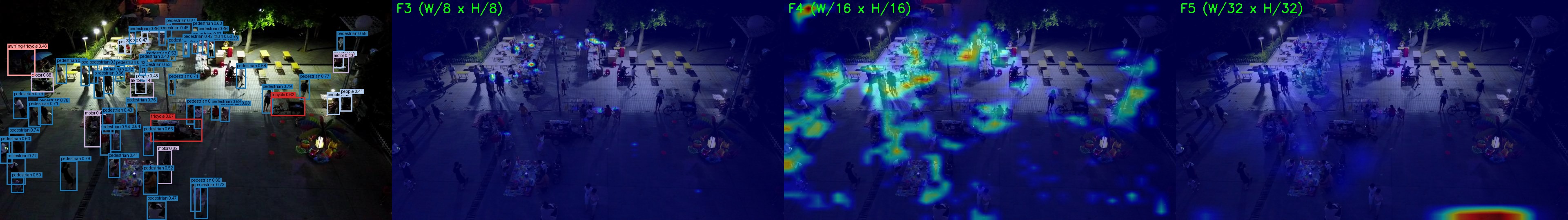}
        \caption{Ours}
        \label{fig:cam_set1_b}
    \end{subfigure}
    
    \caption{\textbf{Grad-CAM Visualizations on VisDrone 2019 val (Set 1).} (a) DEIMv2-X; (b) TinyFormer-X-PBM.}
    \label{fig:supp_vis_visdrone_1}
\end{figure}

\begin{figure}[h]
    \centering
    \begin{subfigure}{\linewidth}
        \centering
        \includegraphics[width=.95\linewidth]{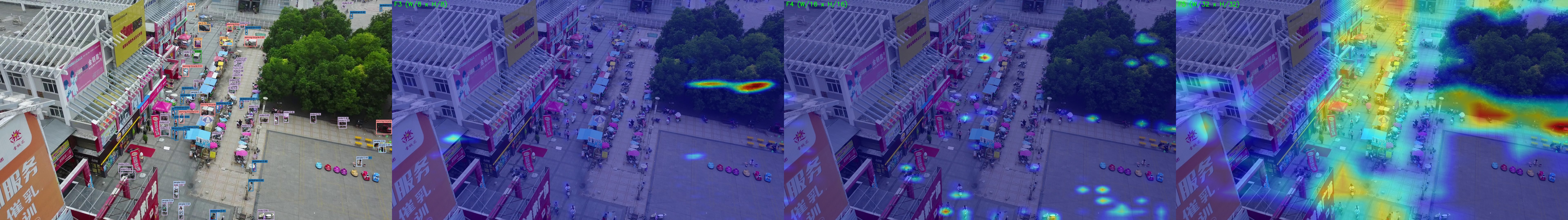}
        \caption{DEIMv2}
        \label{fig:cam_set2_a}
    \end{subfigure} \\
    \vspace{2mm}
    \begin{subfigure}{\linewidth}
        \centering
        \includegraphics[width=.95\linewidth]{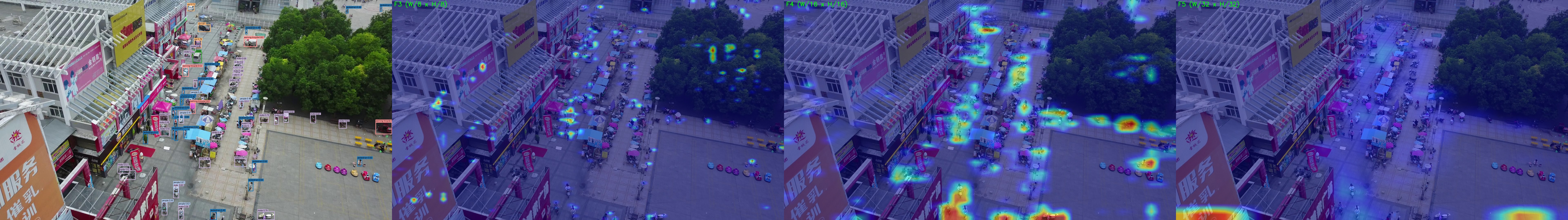}
        \caption{Ours}
        \label{fig:cam_set2_b}
    \end{subfigure}
    
    \caption{\textbf{Grad-CAM Visualizations on VisDrone 2019 val (Set 2).} (a) DEIMv2-X; (b) TinyFormer-X-PBM.}
    \label{fig:supp_vis_visdrone_2}
\end{figure}

\begin{figure}[h]
    \centering
    \begin{subfigure}{\linewidth}
        \centering
        \includegraphics[width=.98\linewidth]{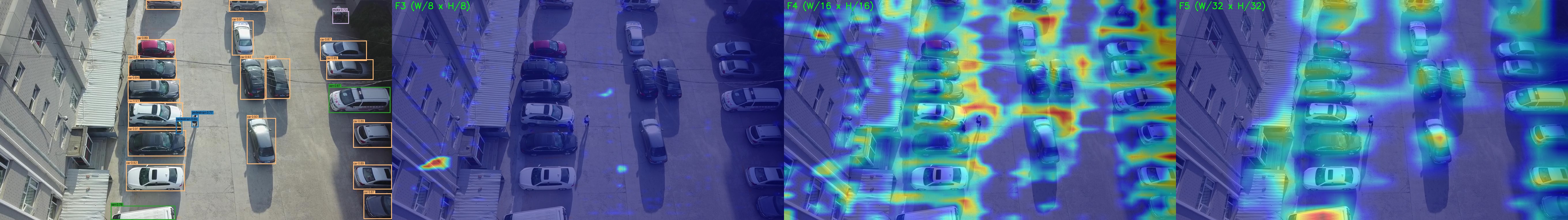}
        \caption{DEIMv2}
        \label{fig:cam_set3_a}
    \end{subfigure} \\
    \vspace{2mm}
    \begin{subfigure}{\linewidth}
        \centering
        \includegraphics[width=.98\linewidth]{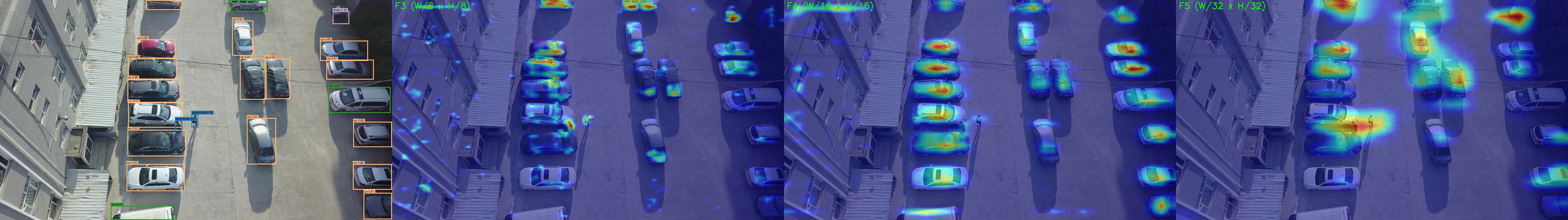}
        \caption{Ours}
        \label{fig:cam_set3_b}
    \end{subfigure}
    
    \caption{\textbf{Grad-CAM Visualizations on VisDrone 2019 val (Set 3).} (a) DEIMv2-X; (b) TinyFormer-X-PBM.}
    \label{fig:supp_vis_visdrone_3}
\end{figure}

\clearpage 

\subsection{Grad-CAM Visualizations on MS COCO val2017}

To evaluate general object detection, we provide MS COCO val2017 visualizations following the same configurations as defined in Sec.~\ref{subsec:visdrone_cam}. These results confirm that TinyFormer maintains stable feature representations across common categories, demonstrating that our enhancements do not compromise general detection capabilities on standard benchmarks.

\begin{figure}[H]
    \centering
    \begin{subfigure}{\linewidth}
        \centering
        \includegraphics[width=.86\linewidth]{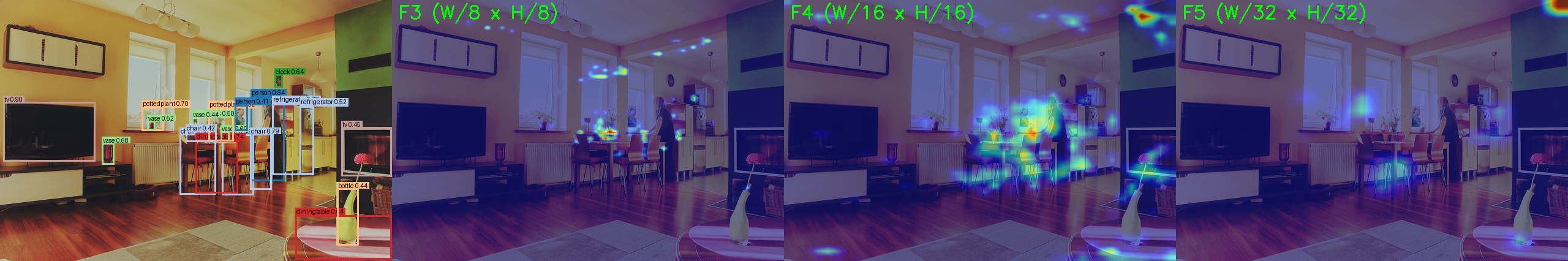}
        \caption{DEIMv2}
        \label{fig:cam_coco1_a}
    \end{subfigure} \\
    \vspace{1mm} 
    \begin{subfigure}{\linewidth}
        \centering
        \includegraphics[width=.86\linewidth]{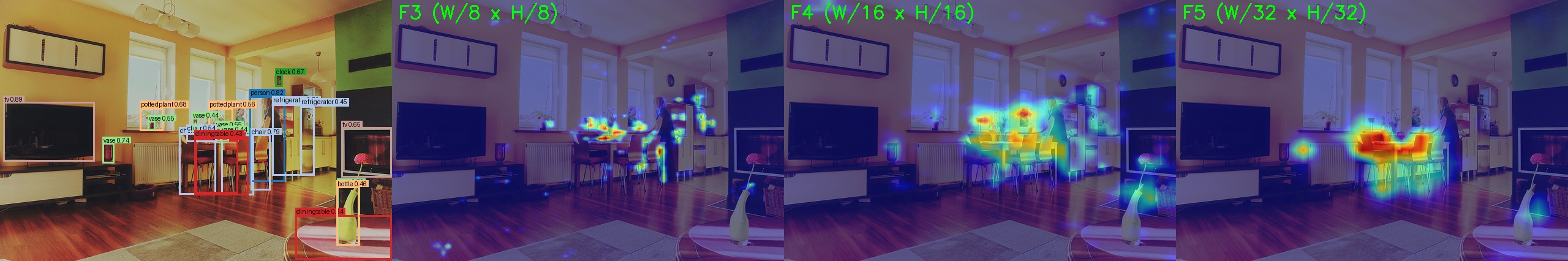}
        \caption{Ours}
        \label{fig:cam_coco1_b}
    \end{subfigure}
    
    \caption{\textbf{Grad-CAM Visualizations on COCO val2017 (Set 1).} (a) DEIMv2-X; (b) TinyFormer-X-PBM.}
    \label{fig:supp_vis_coco_1}
\end{figure}

\vspace{-2mm} 

\begin{figure}[H]
    \centering
    \begin{subfigure}{\linewidth}
        \centering
        \includegraphics[width=.86\linewidth]{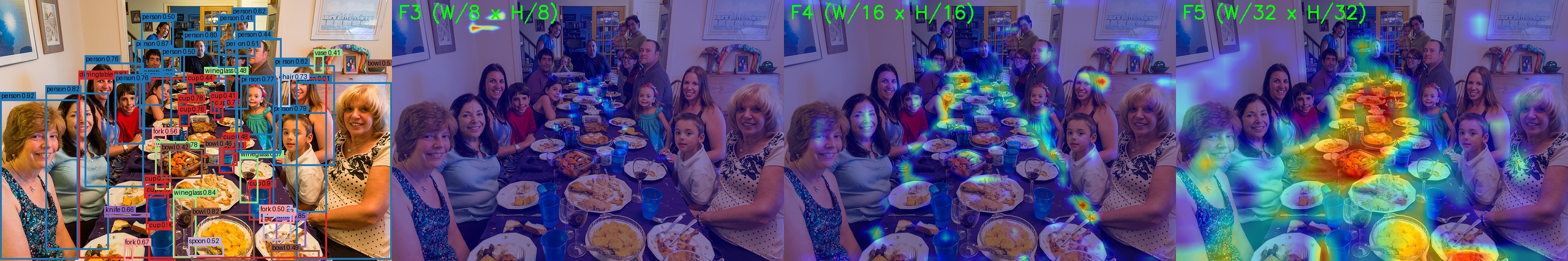}
        \caption{DEIMv2}
        \label{fig:cam_coco2_a}
    \end{subfigure} \\
    \vspace{1mm}
    \begin{subfigure}{\linewidth}
        \centering
        \includegraphics[width=.86\linewidth]{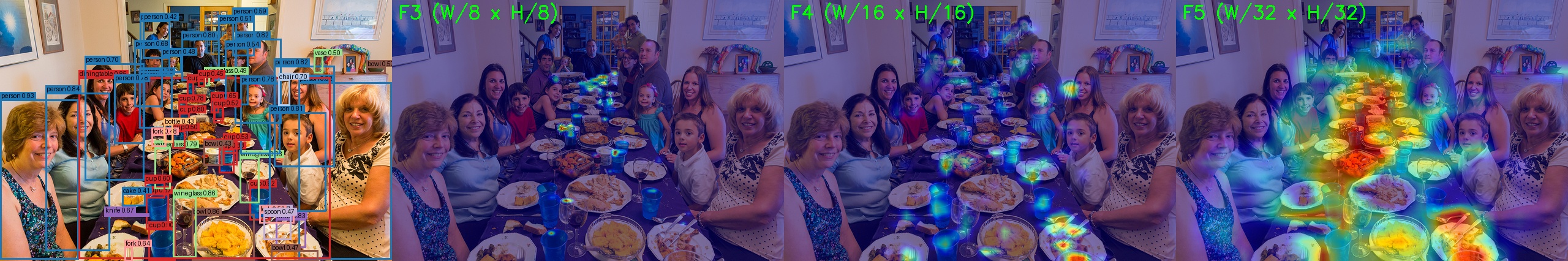}
        \caption{Ours}
        \label{fig:cam_coco2_b}
    \end{subfigure}
    
    \caption{\textbf{Grad-CAM Visualizations on COCO val2017 (Set 2).} (a) DEIMv2-X; (b) TinyFormer-X-PBM.}
    \label{fig:supp_vis_coco_2}
\end{figure}

\vspace{-2mm}

\begin{figure}[H]
    \centering
    \begin{subfigure}{\linewidth}
        \centering
        \includegraphics[width=.86\linewidth]{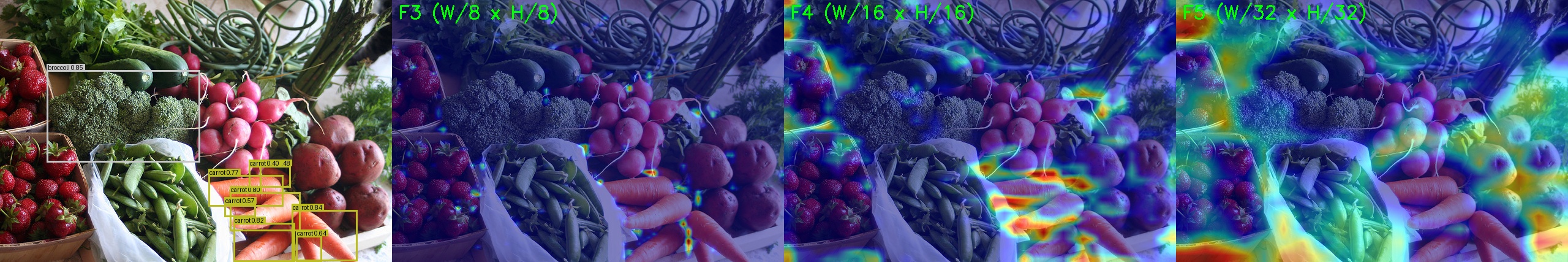}
        \caption{DEIMv2}
        \label{fig:cam_coco3_a}
    \end{subfigure} \\
    \vspace{1mm}
    \begin{subfigure}{\linewidth}
        \centering
        \includegraphics[width=.86\linewidth]{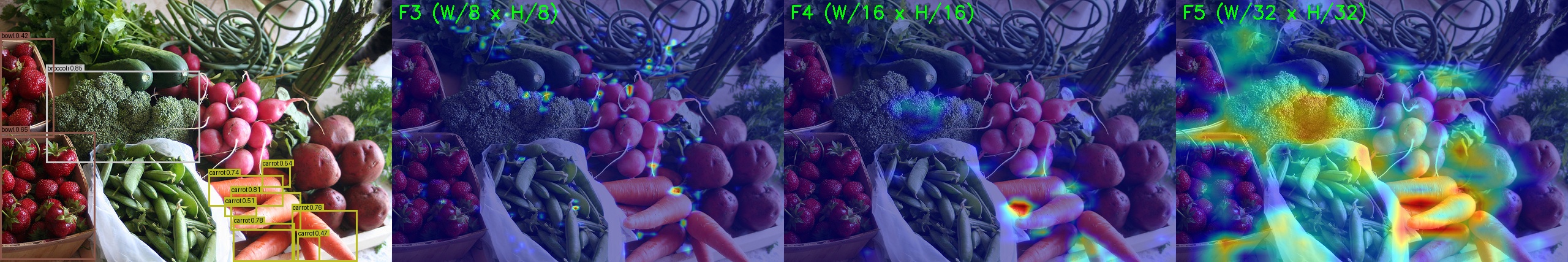}
        \caption{Ours}
        \label{fig:cam_coco3_b}
    \end{subfigure}
    
    \caption{\textbf{Grad-CAM Visualizations on COCO val2017 (Set 3).} (a) DEIMv2-X; (b) TinyFormer-X-PBM.}
    \label{fig:supp_vis_coco_3}
\end{figure}

\section{Design Choices of SSA}

\begin{figure}[h] 
    \centering
    \includegraphics[width=\textwidth]{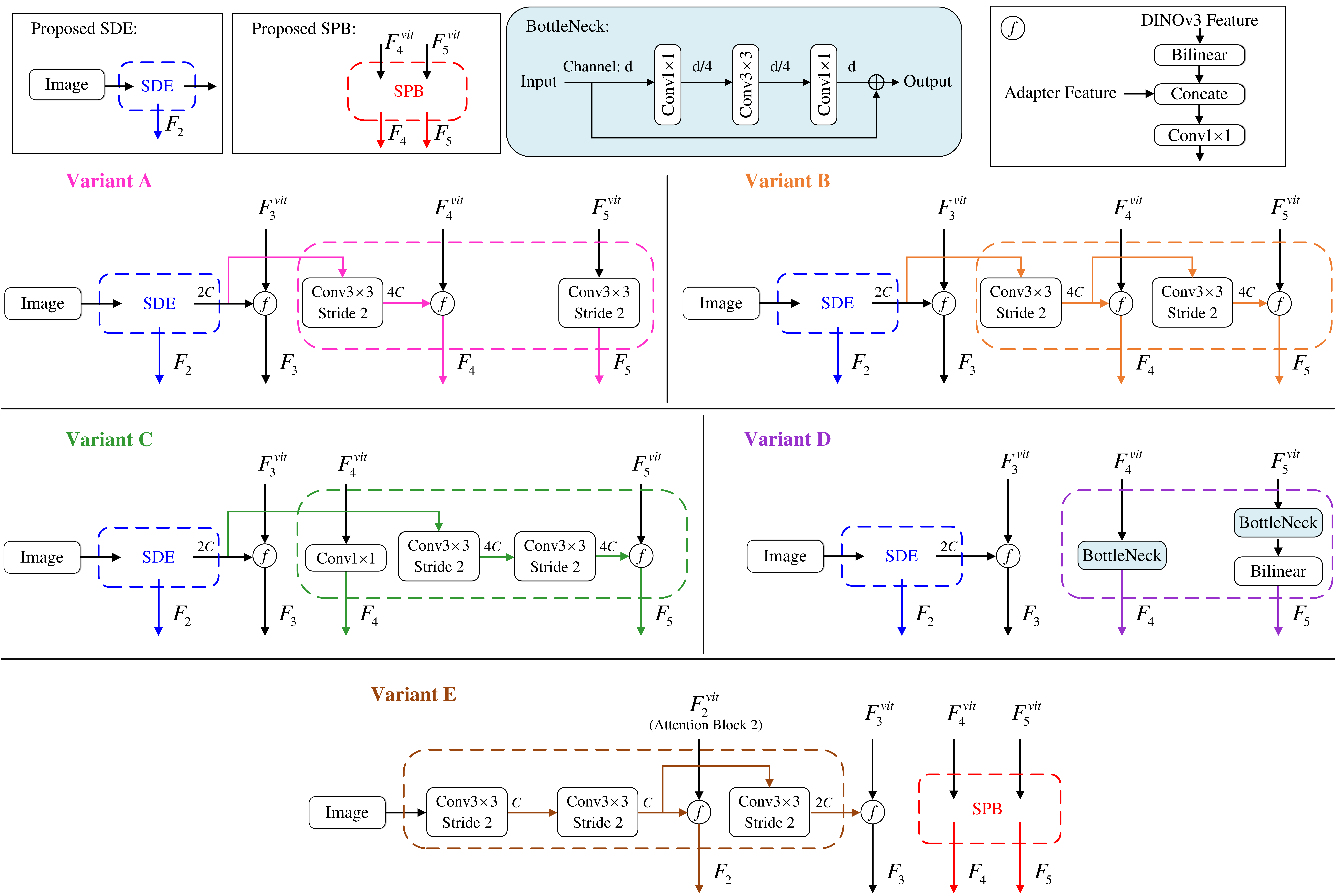} 
    \caption{\textbf{Illustration of the architectural variants of the Spatial Semantic Adapter (SSA).} (Top) Detailed internal structures of the proposed SDE and SPB components, alongside an optional Bottleneck block. (Middle and Bottom) We evaluate five architectural variants (A--E) to determine the optimal configuration for tiny object detection: 
    \textbf{Variant A, B, and C} explore different injection depths, extending spatial cues up to $F_4$, $F_5$, or specific hierarchical combinations; 
    \textbf{Variant D} investigates the impact of replacing standard convolutions with Bottleneck blocks within the SPB to balance complexity and performance; 
    \textbf{Variant E} evaluates the effect of early feature fusion at the $1/4$ scale ($F_2$), where spatial features are integrated with intermediate transformer representations. 
    The notations $C, 2C, 4C$ denote hierarchical channel scaling, and $d$ represents the hidden dimension of the Bottleneck block.}
    \label{fig:ssa_exp}
\end{figure}

Based on the full TinyFormer-X-PBM framework, we investigate the optimal architectural design for SSA to determine where CNN-derived spatial priors best harmonize with ViT-based semantic tokens. As presented in Table~\ref{tab:ablation_ssa_variants} and Fig.~\ref{fig:ssa_exp}, we evaluated our proposed configuration against several variants focusing on injection depth (Variants A--C), module complexity (Variant D), and fusion strategy (Variant E). This investigation is critical because the integration of high-resolution spatial priors into a non-hierarchical Transformer backbone must balance pixel-level precision with the model's capacity for global semantic abstraction.

\begin{table}[h]
\centering
\caption{\textbf{Ablation on different architectural variants of SSA  on MS COCO dataset.} We explore various structural designs, including different injection stages and module types. All models are based on TinyFormer-X-PBM. Bold indicates the optimal overall performance.}
\label{tab:ablation_ssa_variants}
\renewcommand{\arraystretch}{1.1}
\setlength{\tabcolsep}{3pt} 
\begin{tabular}{l | cc | cccc}
\toprule
\textbf{Configuration} & \textbf{Param(M)} & \textbf{FLOPs(G)} & \textbf{AP} & \textbf{AP$_S$} & \textbf{AP$_M$} & \textbf{AP$_L$} \\
\midrule
Variant A (Up to $F_4$) & 51.8 & 165.3 & 58.12 & 39.68 & 62.76 & 75.61 \\
Variant B (Up to $F_5$) & 52.5 & 165.9 & 58.13 & 40.15 & 62.84 & 75.59 \\
Variant C ($F_2$ \& $F_3$ \& $F_5$) & 52.4 & 165.6 & 58.29 & 40.50 & 63.11 & 76.15 \\
Variant D (Bottleneck SPB) & 51.8  & 165.5 & 58.26 & 40.29 & 63.22 & 75.54 \\
Variant E (Early $F_2$ Fusion) & 51.6 & 169.2 & 58.31 & 40.72 & \textbf{63.40} & 75.93 \\
\rowcolor{Gray}
\textbf{Proposed SSA ($F_2$ \& $F_3$)} & 51.5 & 164.2 & \textbf{58.50} & \textbf{40.94} & 63.15 & \textbf{76.55} \\
\bottomrule
\end{tabular}
\end{table}

The results reveal a consistent trend: extending spatial priors into deeper stages does not improve performance and can even degrade detection accuracy for small objects. This suggests that spatial information becomes less beneficial once high-level semantic abstraction has been established. For instance, Variant B, which pushes spatial priors all the way to the $1/32$ scale ($F_5$), results in a noticeable drop in $AP_S$ from $40.94\%$ to $40.15\%$. Similarly, Variant D introduces Bottleneck blocks into the SPB to increase model capacity; however, it yields inferior $AP_S$ ($40.21\%$) while increasing both parameters and FLOPs. We observe that while deeper injection or more complex modules add more pathways, they also introduce additional computational overhead without commensurate accuracy gains for tiny instances.

Furthermore, we explore the effect of early feature fusion in Variant E, where spatial features are integrated with intermediate transformer representations at the $F_2$ stage. Although Variant E achieves a slightly higher $AP_M$ ($63.40\%$), it incurs a significant computational penalty ($169.2$G FLOPs) and fails to match the proposed SSA in $AP_S$ ($40.72\%$ vs. $40.94\%$). These results confirm that maintaining $F_2$ as a pure spatial prior is more effective for tiny object localization than premature interaction with semantic features. Consequently, our findings confirm that restricting the SSA strictly to the early stages ($F_2, F_3$) achieves the optimal architectural balance: it delivers sufficient spatial cues for small instances while maintaining the maximum accuracy and efficiency of the model.

\section{Structural Variations of PBM} 

We further optimize the internal logic of the PBM by ablating its block deployment and fusion operations, as detailed in Table~\ref{tab:ablation_prbm}. 

\textbf{Number of Blocks:} We first assess the impact of hierarchical coverage. Compared to the baseline without PBM, deploying a single bi-fusion block at the intermediate $F_3$ level yields a solid improvement of +0.59\% in $AP_S$. However, the most substantial gain is achieved by deploying two blocks to cover both $F_3$ and $F_4$ scales, resulting in a +1.61\% boost to 40.94\% $AP_S$. This demonstrates that small object localization benefits from a multi-stage spatial reconstruction process that captures cues across varying receptive fields, ensuring that fine-grained details are propagated through the pyramid neck. 

\textbf{Fusion Operations:} A critical aspect of our bi-fusion design is the "Align-then-Injection" protocol. In our proposed configuration, we treat the deep semantic map as a context residual (element-wise addition) while treating the high-resolution shallow map as independent spatial evidence (concatenation). To test this hypothesis, we evaluated a swapped configuration where shallow features were added and deep features were concatenated. This modification led to a noticeable decline in $AP_S$ from 40.94\% to 40.51\%.

\begin{table}[!ht] 
\centering
\caption{\textbf{Ablation on PBM structural designs  on MS COCO dataset.} We test the impact of the number of PBM blocks and the choice of fusion operations. The baseline (0 blocks) corresponds to our base TinyFormer-X model.}
\label{tab:ablation_prbm}

\renewcommand{\arraystretch}{1.1}
\setlength{\tabcolsep}{1.2pt} 

\begin{tabular}{c | l | cc | cccc}
\toprule
\textbf{\# Blocks} & \textbf{Configuration (Scale)} & \textbf{Param(M)} & \textbf{FLOPs(G)} & \textbf{AP} & \textbf{AP$_S$} & \textbf{AP$_M$} & \textbf{AP$_L$} \\
\midrule
0 & Baseline (No PBM) & 49.8 & 151.1 & 58.38 & 39.33 & 63.14 & 76.46 \\
1 & Add $F_{i+1}$, Concate $F_{i-1}$ ($i=3$) & 50.8 & 161.9 & 58.46 & 39.92 & \textbf{63.15} & 75.70 \\
2 & Add $F_{i-1}$, Concate $F_{i+1}$ ($i=3, 4$) & 51.5 & 164.2 & 58.38 & 40.51 & 62.88 & 76.23 \\
\rowcolor{Gray}
\textbf{2} & \textbf{Add $F_{i+1}$, Concate $F_{i-1}$ ($i=3, 4$)} & 51.5 & 164.2 & \textbf{58.50} & \textbf{40.94} & \textbf{63.15} & \textbf{76.55} \\
\bottomrule
\end{tabular}
\end{table}




\clearpage
\end{document}